\documentclass[letterpaper, 10 pt, conference]{ieeeconf}
\IEEEoverridecommandlockouts
\usepackage{cite}
\usepackage{amsmath,amssymb,amsfonts}
\usepackage{algorithmic}
\usepackage{graphicx}
\usepackage{textcomp}
\usepackage{xcolor}
\usepackage{hyperref}
\usepackage{caption}
\usepackage{subcaption}
\usepackage{tablefootnote}
\usepackage{makecell}

\def\myheight{2.3}

\def\BibTeX{{\rm B\kern-.05em{\sc i\kern-.025em b}\kern-.08em
    T\kern-.1667em\lower.7ex\hbox{E}\kern-.125emX}}

\title{\LARGE \bf
LVLane: Deep Learning for Lane Detection and Classification in Challenging Conditions}

\author{Zillur Rahman and Brendan Tran Morris% <-this % stops a space
\thanks{Zillur Rahman and Brendan Tran Morris are with the Department of Electrical and Computer Engineering, University of Nevada, Las Vegas, USA.
{\tt\small zillur991@gmail.com},  {\tt\small brendan.morris@unlv.edu}}%
}

\begin{document}

\maketitle
\thispagestyle{empty}
\pagestyle{empty}

\begin{abstract}
Lane detection plays a pivotal role in the field of autonomous vehicles and advanced driving assistant systems (ADAS). Despite advances from image processing to deep learning-based models, algorithm performance is highly dependent on training data matching the local challenges   %Over the years, numerous algorithms have emerged, spanning from rudimentary image processing techniques to sophisticated deep neural networks. The performance of deep learning-based models is highly dependent on the quality of their training data. Consequently, these models often experience a decline in performance when confronted with challenging scenarios 
such as extreme lighting conditions, partially visible lane markings, and sparse lane markings like Botts' dots. 
To address this, we present an end-to-end lane detection and classification system based on deep learning methodologies. In our study, we introduce a unique dataset meticulously curated to encompass scenarios that pose significant challenges for state-of-the-art (SOTA) lane localization models. %Through fine-tuning selected models, we aim to achieve enhanced localization accuracy. 
Moreover, we propose a CNN-based classification branch, seamlessly integrated with the detector, facilitating the identification of distinct lane types. This architecture enables informed lane-changing decisions and empowers more resilient ADAS capabilities. We also investigate the effect of using mixed precision training and testing on different models and batch sizes. Experimental evaluations conducted on the widely-used TuSimple dataset, Caltech Lane dataset, and our LVLane dataset demonstrate the effectiveness of our model in accurately detecting and classifying lanes amidst challenging scenarios. Our method achieves state-of-the-art classification results on the TuSimple dataset. The code of the work can be found on \url{www.github.com/zillur-av/LVLane}.
\end{abstract}

\section{Introduction}
In recent years, remarkable advancements in autonomous vehicles and intelligent transportation systems have revolutionized how we envision transportation's future. Central to the development of such systems is the capability to accurately detect and interpret lanes on the road, enabling vehicles to navigate safely and efficiently. Lane detection algorithms, a critical component of these systems, allow vehicles to perceive and understand the road environment, making them an indispensable part of advanced driving assistance systems (ADAS) and autonomous driving (AD).
%
%Lane detection techniques have garnered significant attention due to their vital role in enhancing road safety and enabling autonomous driving. 
The detection of lane markings, such as lane boundaries and road center lines, helps vehicles establish their position within the road infrastructure, allowing for precise trajectory planning and reliable navigation. By continuously monitoring the road ahead, lane detection systems facilitate lane departure warning, lane-keeping assistance, and autonomous lane following, which are core functionalities of ADAS \cite{bar2014recent}.
\begin{figure}[b]
     \begin{subfigure}[t]{0.24\textwidth}
         \includegraphics[width=4.25cm, height=3cm]{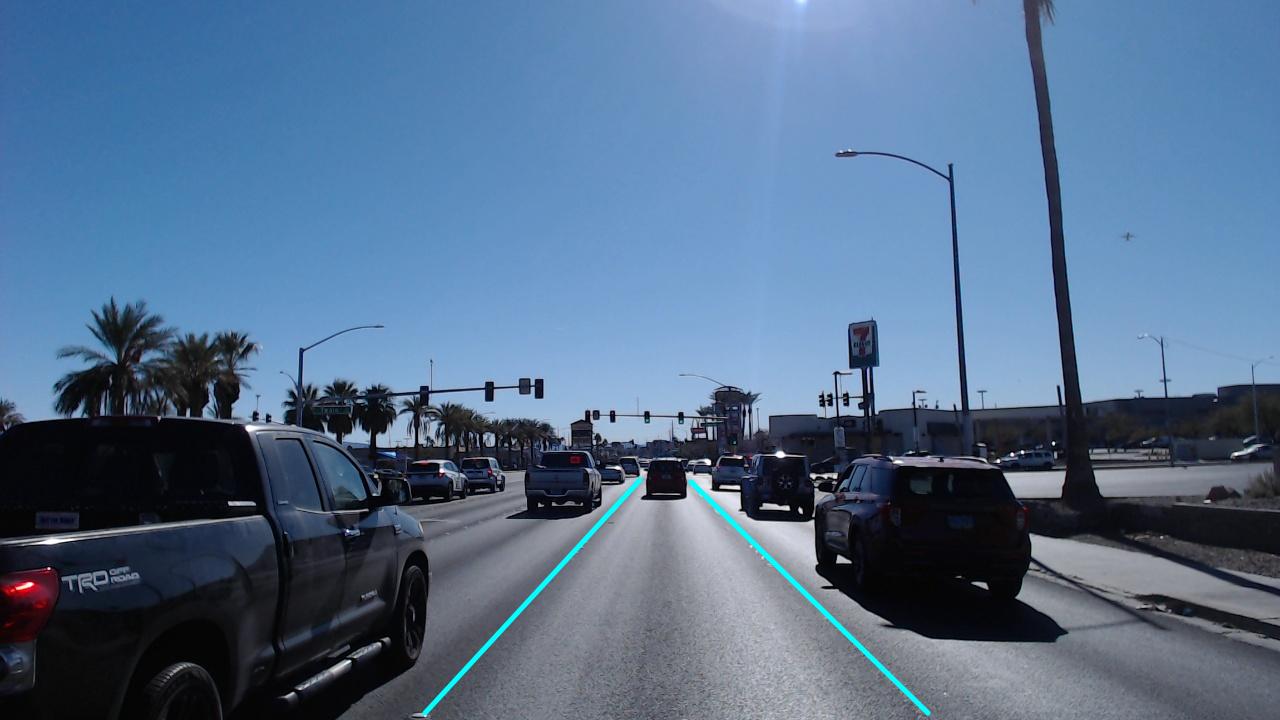}
         \caption{}
         \label{fig: challenging-images (a)}
     \end{subfigure}
     \begin{subfigure}[t]{0.24\textwidth}
         \includegraphics[width=4.25cm, height=3cm]{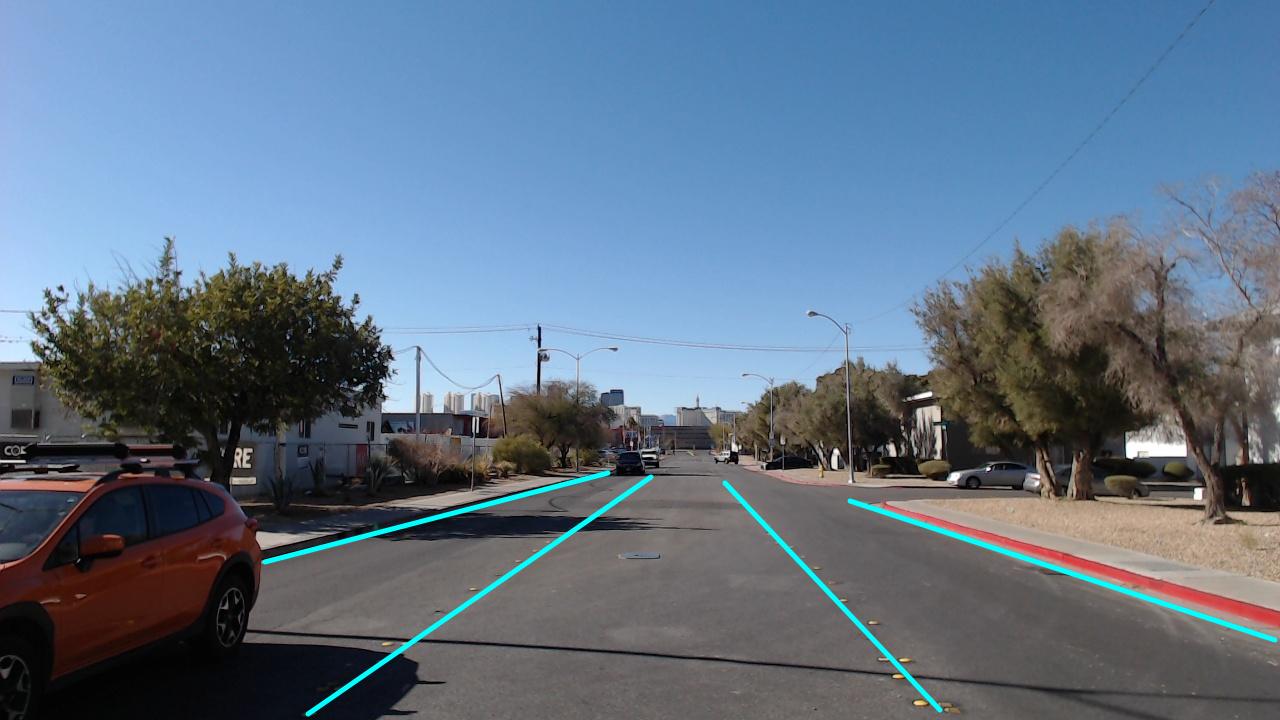}
         \caption{}
        \label{fig: challenging-images (b)}
     \end{subfigure}
        \caption{Challenging images from Las Vegas streets with ground truth (cyan lines). (\subref{fig: challenging-images (a)}) Poor visibility due to road reflections in harsh lighting conditions. (\subref{fig: challenging-images (b)}) Large spacing between Botts' dots.}
        \label{fig: challenging-images}
\end{figure}

Lane detection poses several challenges due to road environments' complex and dynamic nature. Varying lighting conditions such as shadows, bright sunlight, and nighttime illumination can affect the visibility and contrast of lane markings as can be seen from Fig. \ref{fig: challenging-images (a)}. Moreover, lane markings can be partially or fully occluded by other vehicles, pedestrians, or environmental factors like rain or fog. Roads also come in different shapes and configurations, including straight, curved, and intersections. Besides, lane markings can have diverse appearances depending on the region, road type, and maintenance level (e.g. the Botts' dots shown in Fig. \ref{fig: challenging-images (b)} which are prevalent in Western USA). These variations can include different colors, widths, and patterns. %Lane detection algorithms must be adaptable to handle such variations and generalize well across different road environments. 
Road surfaces may contain various visual noises, such as cracks, shadows, or tire marks, which can be mistakenly detected as lane markings. Lane detection algorithms should be adaptable to variation, resilient to noise, and capable of distinguishing between true lane markings and irrelevant visual elements across different road environments.

%Previously, image processing-based techniques were used to address these challenges \cite{aly2008real, bertozzi1998gold}. However, those methods do not work well with lighting variations and occluded scenes. Recently, Deep Learning and  Convolutional Neural Networks (CNN) \cite{he2016deep, simonyan2014very} have had great success in the computer vision field because of their learning ability. The hierarchical nature of CNNs allows them to capture both low-level and high-level visual information, enabling robust and accurate analysis of complex visual data. As a result, we are witnessing applications \cite{rahman2020real} of deep-learning computer vision algorithms in numerous sectors.

\begin{figure}[!t]
    \centering
    \begin{subfigure}{0.5\textwidth}
         \includegraphics[width=0.48\linewidth]
         {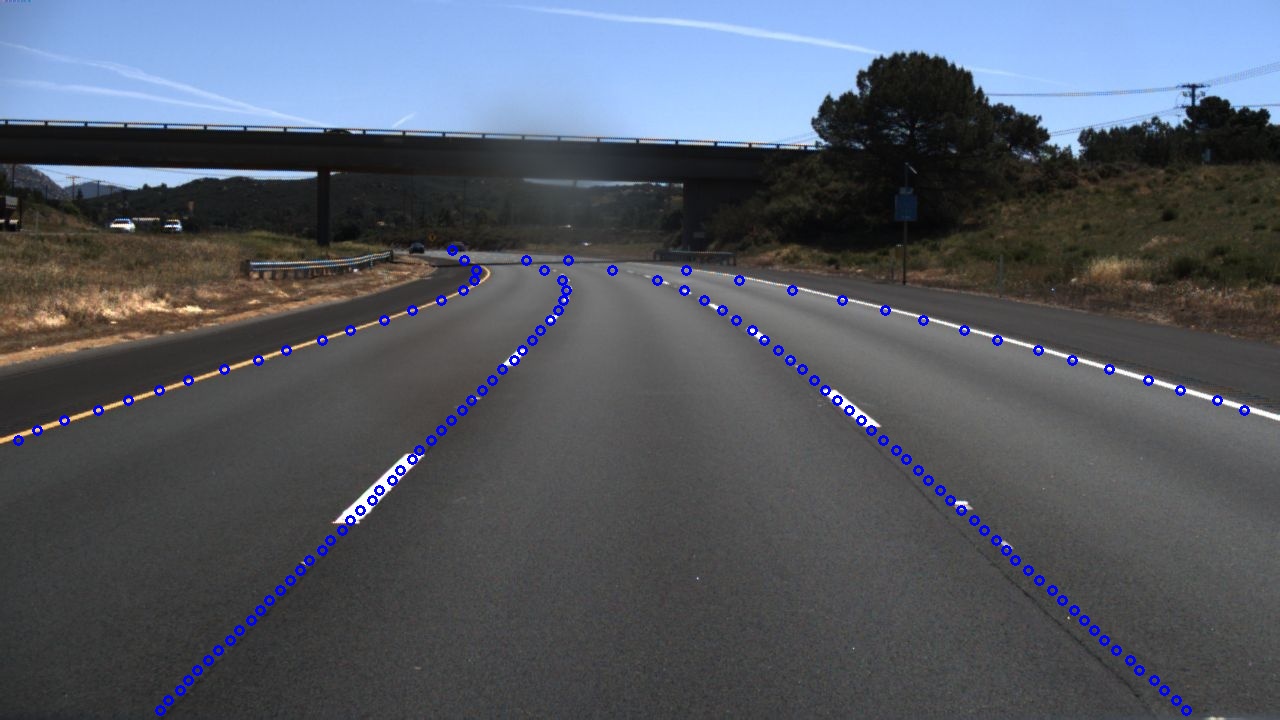}
         \includegraphics[width=0.48\linewidth]{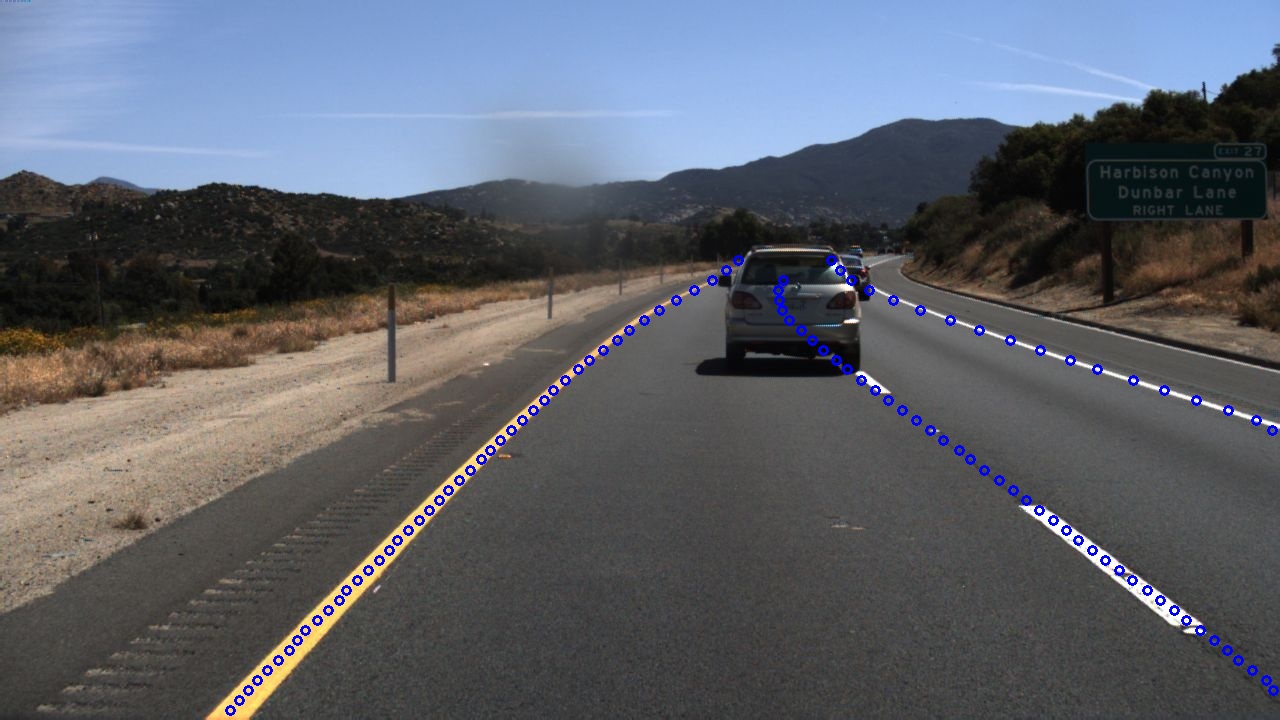}
         \caption{UFLD trained and tested on TuSimple Highway}
         \label{fig:tusimple_highway}
    \end{subfigure}
    \begin{subfigure}{0.5\textwidth}
         \includegraphics[width=0.48\linewidth]{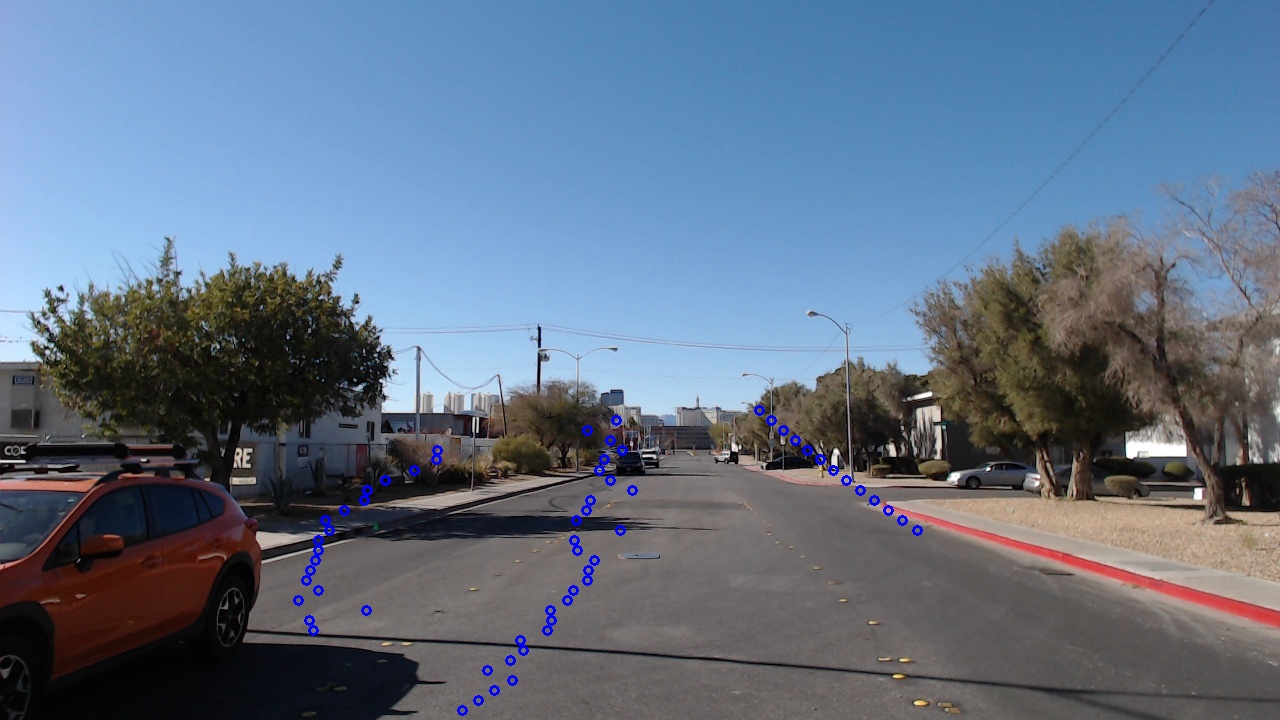}
         \includegraphics[width=0.48\linewidth]{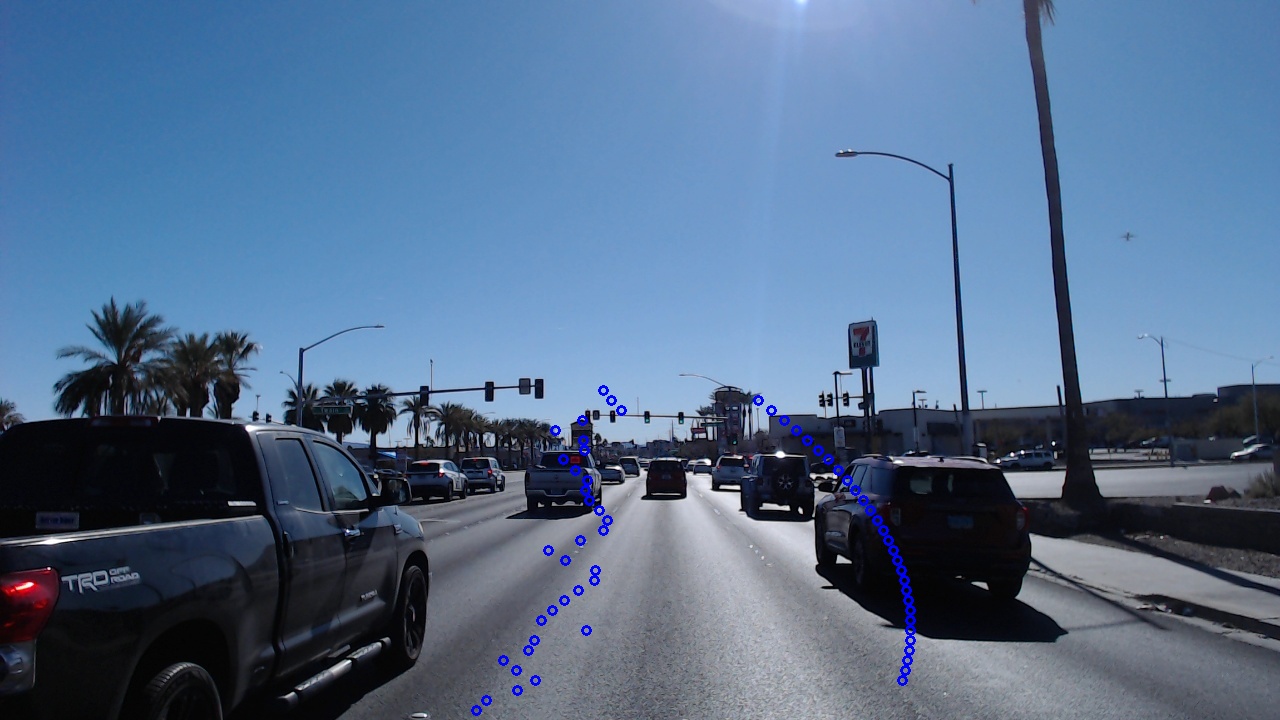}
         \caption{UFLD trained on TuSimple and tested Las Vegas Lanes}
         \label{fig:lasvegaslanes-tusimple}
    \end{subfigure}
    \begin{subfigure}{0.5\textwidth}
         \includegraphics[width=0.48\linewidth]{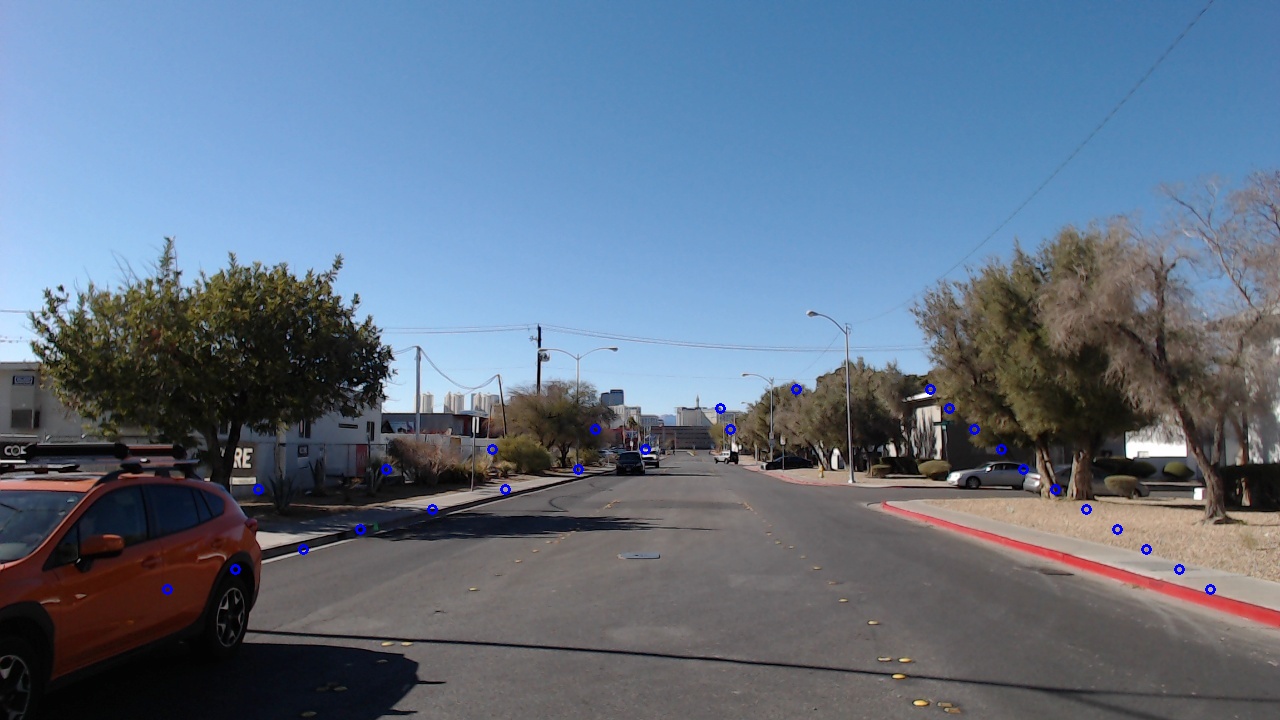}
         \includegraphics[width=0.48\linewidth]{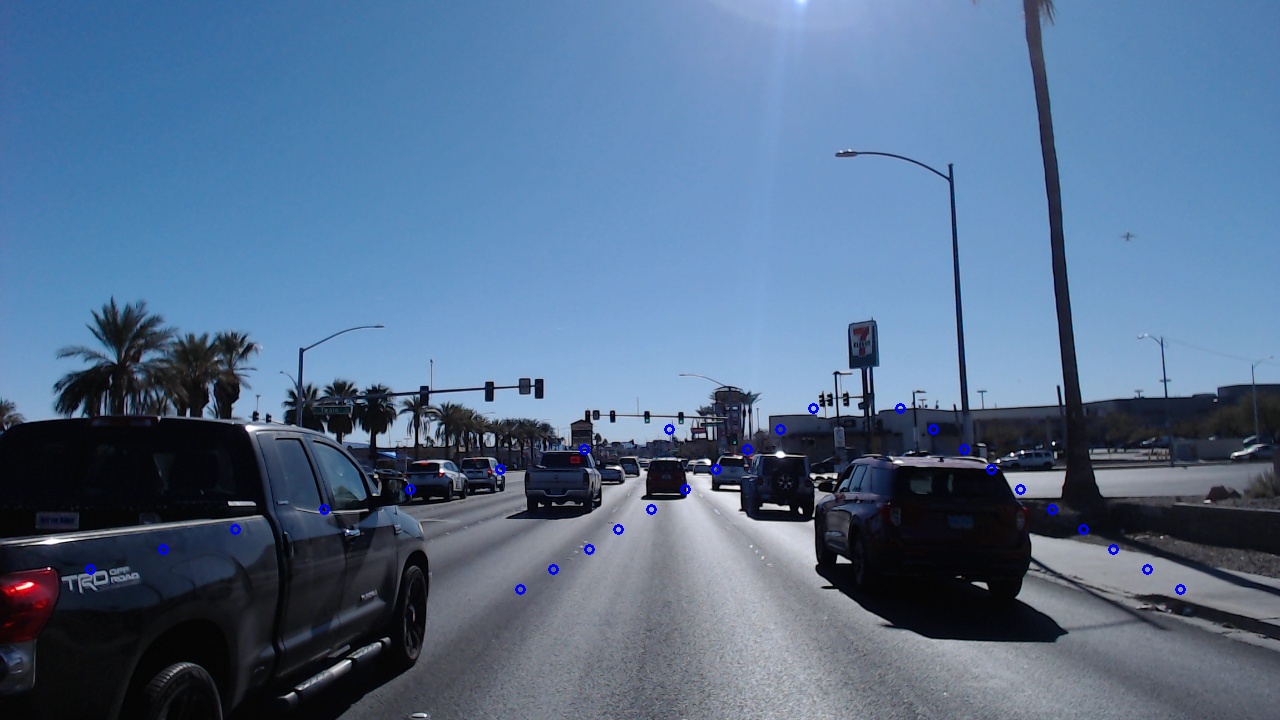}
         \caption{UFLD trained on CuLane and tested Las Vegas Lanes}
         \label{fig:lasvegaslanes-culane}
    \end{subfigure}
    \caption{Strong SOTA lane performance on (\subref{fig:tusimple_highway}) typical highway scene versus poor detection in (\subref{fig:lasvegaslanes-tusimple} and \subref{fig:lasvegaslanes-culane}) the challenges of Las Vegas despite using popular lane datasets.}
    \label{fig:lane_sota}
\end{figure}

Lane detection has been a well-researched area in computer vision and AD, with recent deep convolutional algorithms and techniques such as \cite{pan2018spatial, qin2020ultra, zheng2021resa, zheng2022clrnet} developed to accurately detect lane markings on roads. However, those highly accurate models do not always work well in unseen environments. When lane geometry and marking type become different than the trained model datasets, even state-of-the-art (SOTA) models provide poor performance. As we can see in Fig. \ref{fig:lane_sota} %\ref{fig:tusimple_highway}, 
the SOTA model UFLD\cite{qin2020ultra} works well in an image from the popular TuSimple \cite{tusimple} dataset. On the other hand, the same architecture trained on TuSimple or CuLane\cite{pan2018spatial} (large arterial dataset) shows poor results on a new image from Las Vegas streets.% as shown in Fig. \ref{fig:lasvegaslanes-tusimple} and \ref{fig:lasvegaslanes-culane}.

Furthermore, compared to lane detection, lane classification has received significantly less attention in the research community. Lane classification refers to the task of categorizing detected lane markings into different types, such as solid lines, dashed lines, or double lines. Proper lane classification is crucial for understanding road configurations and lane usage, enabling more ADAS and decision-making algorithms. On top of that, to the best of our knowledge, there are no publicly available quality lane classification datasets, besides Caltech \cite{aly2008real}, which makes this area difficult to explore. 

Another issue we encounter while running existing SOTA models is their high computational cost. % caused by their deeper architectures 
 Present AD vehicles and ADAS solutions must process data from various sensors including multiple cameras, LiDAR, and RADAR with limited resources. %Processing all of these sensor data require powerful computational devices. 
As a result, it is imperative to reduce the computational complexity and speed up the inference process of lane detection models.

With the above motivations, we propose an end-to-end deep learning lane detection and classification model. The main contributions of our work can be summarized as follows:
%To solve the performance degradation in challenging Las Vegas streets, we introduce the LVLane dataset consisting of lane marking annotation and lane classes collected from city streets. To address the lane classification task, we introduce a simple classification network branch that can be used in addition to a deep lane detection network. The branch takes a task-specific feature map from the network's top layers and infers the type of lane. It is lightweight and hence adds a low computation requirement. Finally, to make our model train and test fast, we investigate the effect of the mixed precision technique that speeds up the forward and backward passes. To demonstrate the effectiveness of our model, we train and test it on popular TuSimple\cite{tusimple}, Caltech lane dataset\cite{aly2008real}, and our proposed LVLane dataset. The experiments show our model can accurately detect and classify lane markings in challenging scenarios.

\begin{itemize}
  \item We introduce a novel lane detection and classification dataset from challenging Las Vegas streets to improve performance in city areas. The dataset contains curved lanes, highly occluded scenes, and roads with no visible lane markings.
  \item We propose a simple yet effective lane classification network that can be used with a detection model to identify lane types. The entire model is trained and tested in an end-to-end fashion.
  \item We provide lane class annotations for the TuSimple Dataset along with a simple annotation method designed to facilitate the creation of datasets in the TuSimple format. This resource allows for efficient dataset creation and testing, promoting ease and convenience in the annotation process.
  \item To train and test the models faster, we investigate the effect of using the mixed precision technique\cite{micikevicius2017mixed}.
\end{itemize}
To demonstrate the effectiveness of our model, we train and test it on popular TuSimple\cite{tusimple}, Caltech lane dataset\cite{aly2008real}, and our proposed LVLane dataset. The experiments show our model can accurately detect and classify lane markings in challenging scenarios.

\section{Relevant Work}
\subsection{Lane Detection Methods}
Conventional methods typically address the challenge of lane detection by relying on visual cues. These methods primarily leverage image processing techniques, such as utilizing the HSI color model, to exploit and extract valuable information from the visual data\cite{sun2006hsi,wang2000lane}. Deep learning-based methods use neural networks and have had great success in the computer vision field because of their learning ability. The hierarchical nature of CNNs allows them to capture both low-level and high-level visual information, enabling robust and accurate analysis of complex visual data. As a result, we see applications \cite{rahman2020real} of deep learning in numerous sectors. There are mainly two types of deep learning-based lane detection models: segmentation-based and anchor based. Segmentation methods attempt to identify each lane pixel in the image. On the other hand, anchor-based lane detection only searches for a limited number of key features based on predefined anchor points or boxes which improves computation time. As our target is to evaluate SOTA models, trained on TuSimple, on Las Vegas streets, in our literature review, we mainly focus on that.

\subsubsection{Segmentation-Based Methods:} In the deep-learning area, SCNN\cite{pan2018spatial} introduces a message-passing mechanism to tackle the challenge of the absence of visual evidence, effectively capturing the robust spatial relationships among lanes. This approach yields substantial enhancements in lane detection performance, however, the real-time applicability of the method is hindered by its relatively slower processing speed. RESA \cite{zheng2021resa} is a segmentation-based lane detection model with three modules. First, a CNN feature extractor works as an encoder, then the RESA module captures the spatial relationship of pixels across rows and columns. It shifts feature maps in vertical and horizontal directions with different strides. Thus, it can detect lanes in occluded scenarios. Finally, a decoder transforms the low-resolution feature map into full image size for pixel wise prediction. CurveLane-NAS\cite{xu2020curvelane} employs neural architecture search (NAS) to identify an improved network architecture that effectively captures precise information, thereby enhancing curved lane detection. Nevertheless, the computational cost of NAS is exceptionally high, demanding substantial GPU resources and extensive hours of computation.

\subsubsection{Anchor-Based Methods:} UFLD \cite{qin2020ultra} presents a row-anchor-based lane detection method that looks for lanes at some predefined target locations. %It looks for lanes at some predefined target location instead of segmenting every pixel. This makes it faster. 
Also, it proposes a structural loss function that uses prior lane shape, and direction to solve the visual-no-clue problem. In the work presented by CondLaneNet\cite{liu2021condlanenet}, a conditional lane detection strategy is introduced that relies on conditional convolution and a row anchor-based formulation. The approach involves first identifying the start points of lane lines and subsequently performing lane detection based on row anchors. In CLRNet\cite{zheng2022clrnet}, they use a novel network called Cross Layer Refinement Network that utilizes both global and local features. It uses a module called ROIGather to detect lanes with high semantic features and then performs refinement based on low-level features. Additionally, it proposes a LineIoU loss function that further improves localization accuracy.  

\subsection{Lane Classification Methods}
After an extensive literature search, we found only a few lane classification systems based on classical image processing techniques and machine learning algorithms. 

In \cite{de2015automatic}, a two-stage cascaded Bayesian lane classification model is introduced. The maximum and minimum intensity profiles of lane markings are used to extract lane features. Since it depends solely on pixel intensity, the model performs poorly in varying lighting conditions. \cite{song2018lane} suggests a simple Hough transform-based lane detector which is transformed into a region of interest (ROI) image and used as input to the CNN classifier after noise removal. %and a CNN-based lane classifier. The output of the detector is transformed into a region of interest (ROI) image and used as input to the CNN classifier after noise removal. It shows good performance on the KITTI test set\cite{geiger2012we} with substantial ROI post-processing. 
The Cascaded CNN\cite{pizzati2020lane} proposes one instance segmentation model for lane detection and another CNN, which uses the feature descriptor of the detector, for lane classification. %The output of the detector is converted to feature descriptors of size 256x256 and then send to the classifier. 
It achieves satisfactory performance on both tasks for the TuSimple dataset; however, training two networks separately and then combining them makes it complicated to use. In another work \cite{sim}, an unsupervised domain adaptation (UDA) approach is introduced which trains using a synthetic dataset created from the CARLA simulator \cite{dosovitskiy2017carla}.% is used to train a model and test it on real-world images.
The model shows satisfactory performance on the detection part but not on the classification part indicating the gap between simulation and real-world test cases.

\section{Method}
\begin{figure}[t]
\centerline{\includegraphics[width=0.95\linewidth]{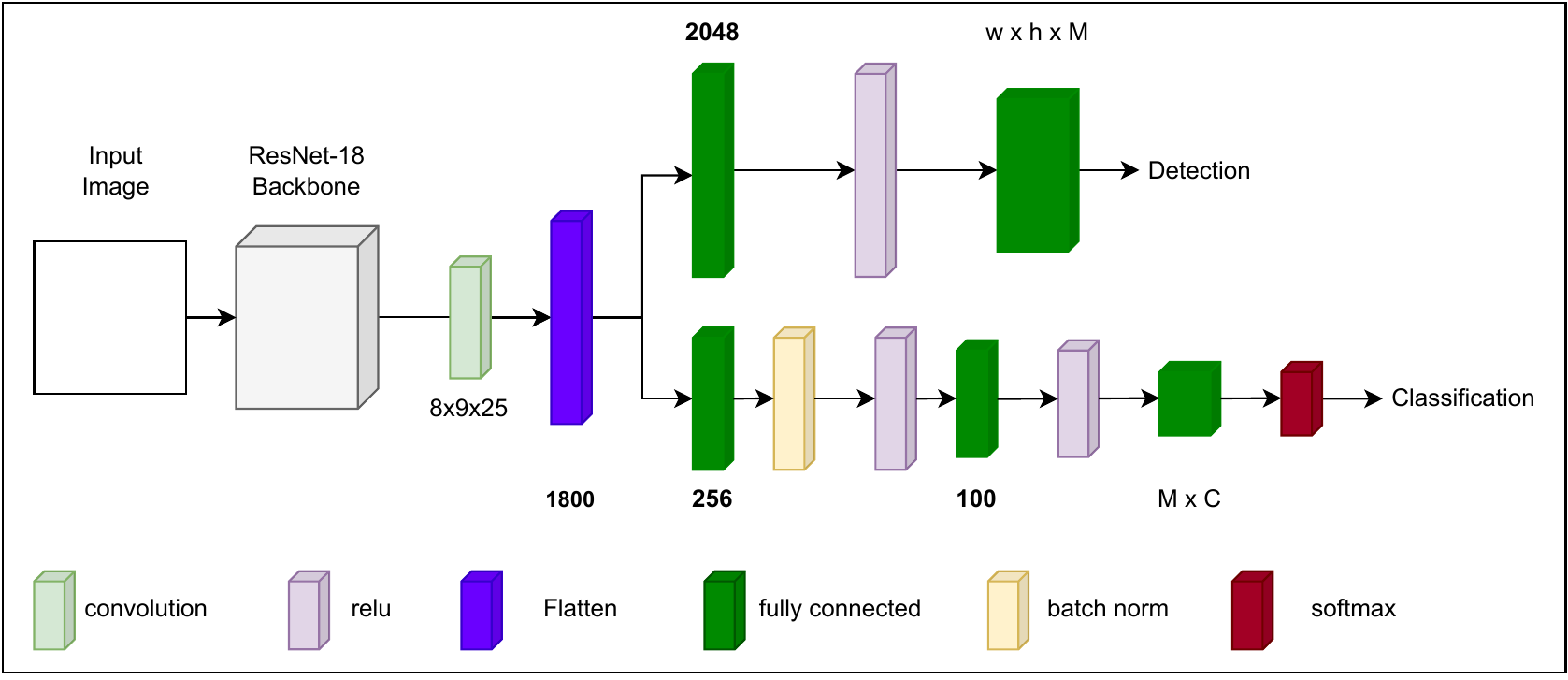}}
\caption{Lane Classification branch with lane detection.}
\label{fig:entire-system}
\end{figure}

Our approach for lane analysis is to fine-tune and extend SOTA detection models using the public TuSimple dataset in addition to a new and challenging urban lane dataset. 
Our entire model (Fig. \ref{fig:entire-system}) has three major modules. The first is the convolutional feature extractor. The second is the lane detector that detects lanes from input images and provides the pixel location of each lane. The third is the lane classifier which uses the feature map from the feature extractor to classify each detected lane type. The network is trained and tested in an end-to-end way.
\subsection{Dataset Collection} 
During our investigation into the performance disparity observed among various SOTA lane detection models depicted in Fig. \ref{fig:lane_sota}, we have identified several crucial insights. The majority of current SOTA models are trained on either the TuSimple or CuLane datasets, each possessing unique characteristics. The TuSimple dataset predominantly comprises highway images with straight lanes, minimal traffic congestion, limited occlusion, and highly visible markings. In contrast, the CuLane dataset offers a more challenging scenario with over 100,000 images encompassing diverse conditions such as night scenes, curves, crowds, and normal environments. However, models based solely on the CuLane dataset struggle to detect complex lane markings prevalent on Las Vegas streets, as illustrated in Fig. \ref{fig:lasvegaslanes-culane}, due to the absence of circular Botts' dots markings in training which are present in TuSimple. As a result, models trained on those datasets do not perform well in bright light conditions, partially visible lane markings, and Botts' dots lanes. 

Therefore, we recognize the necessity for a new dataset encompassing challenging scenarios, which we collected by driving across city areas. Equipped with a Logitech web camera mounted on the front of the vehicle, we continuously captured photos with a resolution of 1280x720. Subsequently, we manually curated a selection of pertinent images, constituting our comprehensive dataset that includes varying weather and lighting conditions.

As the annotation tool, we use VIA \cite{dutta2016via, dutta2019vgg}. %This tool is easy to use for annotating computer vision images. 
Since it does not provide lane detection annotation in TuSimple format, we use polyline, then use Cubic Spline to transform the labels to TuSimple format. Our dataset has a total of 7 classes: solid-yellow, solid-white, dashed, double-dashed, Botts'-dots, double-yellow, and road edge/unknown marking.

%\includesvg[width=10cm]{images/system-overview}

\subsection{Lane Detection}
With our primary objective centered around enhancing lane detection capabilities to excel in Las Vegas scenarios akin to Fig. \ref{fig: challenging-images}, we can capitalize on the exceptional learning capabilities exhibited by SOTA models. %Despite their underperformance on our specific images, the deeper architectures of these models are inherently designed to acquire intricate features from pertinent data sources. 
By employing fine-tuning techniques and leveraging our demanding dataset to focus on acquiring distinguishing features, we anticipate significant performance improvements. In the current stage of our research, we have thoroughly investigated UFLD\cite{qin2020ultra} and introduced modifications to their architectures to align with our specific task requirements. We also report performance measures for RESA \cite{zheng2021resa} to make comparisons in some cases.

We opted for UFLD as our choice due to its lightweight architecture, which achieves a high frame rate of 300+ FPS while maintaining comparable performance to state-of-the-art methods. UFLD tackles the lane detection task using a row-based selection technique, where lanes are represented by a series of horizontal positions at predetermined rows referred to as row anchors. The position of each row anchor is subdivided into a specified number of gridding cells, denoted as ``$w$". Consequently, predicting the position for the ``$i$-th" lane and ``$j$-th" row anchor becomes a classification challenge. The model outputs the probability, $P_{i,j}$, of selecting ($w$ + 1) gridding cells, with the additional dimension in the output denoting the absence of a lane. The lane localization loss is defined as follows:
\begin{equation}
    L_{cls} = \sum_{i=1}^M \sum_{j=1}^h L_{CE} (P_{i,j},T_{i,j})
\end{equation} 
$M$ is the number of lanes, $h$ is the number of row anchors, $T_{i,j}$ is the target, and $L_{CE}$ is the cross entropy loss function. 

The total detection loss is given as the weighted sum of the lane localization loss and two structural loss functions introduced to model the location relation of lane points: 
\begin{equation}
    L_{detection} = L_{cls} + \alpha *L_{sim} + \lambda *L_{shp}.
\end{equation} 
where the loss coefficients $\alpha$ and $\lambda$ could be between 0 and 1.  The similarity loss ($L_{sim}$ is derived from the idea that lane points in adjacent row anchors should be close to each other. The shape loss ($L_{shp}$) is derived from the idea that most of the lanes are straight.  Full details on the structural loss can be found in the UFLD paper \cite{qin2020ultra}.

We use ResNet18\cite{he2016deep} as the backbone of UFLD as this is lightweight and works well considering the smaller network size. Besides, we detect a maximum of 6 lanes instead of 4 lanes suggested in the original UFLD study as our dataset, as well as TuSimple, has many images with more than 4 lanes. Further, we exclude the auxiliary branch included in the UFLD network as it is not necessary for our case.

\subsection{Lane Classification}
The lane classifier is a shallow network that works along with the lane detector. %The overview of the final whole model is shown in Fig. \ref{fig:entire-system}. We design a shallow network as the classification branch. 
The deep neural network architecture starts with an input image, which is then passed through a ResNet18 feature extractor. The output from ResNet18 is fed into a convolutional layer, which further processes the features. After the convolutional layer, the feature map is flattened to convert it into a vector representation. At this point, the network splits into two branches: one for lane detection and the other for classification. 

%In the classification branch, the flattened feature vector is input to a sequence of layers designed to learn and extract relevant features for classification. These layers typically include a dense layer followed by batch normalization and a rectified linear unit (ReLU) activation function. The dense layer performs a linear transformation on the input features, followed by the batch normalization layer that normalizes the outputs and improves the stability of the network. The ReLU activation function introduces non-linearity, allowing the network to learn complex relationships in the data.

In the classification branch, the flattened feature vector is passed into a series of fully connected layers. We use one Dense+Batch-Norm+ReLU sequence and another Dense+ReLU sequence. Finally, the last dense layer is employed with a softmax activation function, which produces the final classification probabilities across $C$ different classes.

\subsection{Mixed Precision Technique}
The automatic mixed precision process\cite{micikevicius2017mixed} is a technique used to accelerate the training and inference of deep learning models by leveraging the advantages of mixed precision arithmetic. In mixed-precision training, weights are transformed into lower-precision (FP16) to expedite computations. Gradients are then computed using the lower-precision weights, converted back to higher-precision (FP32) to ensure numerical stability, and subsequently used to update the original weights with scaled gradients. This process optimizes the training procedure by striking a balance between computational efficiency and numerical precision.

Deep learning models can benefit from faster training and inference times without sacrificing significant accuracy by employing automatic mixed precision. The reduced precision operations lead to increased computational efficiency, making it especially valuable for large-scale models and resource-constrained environments such as edge devices or cloud-based deployments.

\section{Experiments}
\subsection{Datasets}
The experiments are conducted with three different datasets -- TuSimple, Caltech Lanes, and LVLane datasets. Unless noted otherwise, models were pre-trained on TuSimple. The details of the datasets are shown in Table \ref{tab: datasets}.

The TuSimple dataset was collected on highways under stable lighting conditions. There are 3268 images in the training set, 358 in the validation set, and 2782 images in the test set. The size of the images is 1280x720. The lane types of this dataset were not provided but \cite{pizzati2020lane} annotated the train and validation sets to 7 types of lane classes. We annotate the remaining test set.

The Caltech Lanes dataset comprises four video clips captured at various times of the day on streets in Pasadena, CA. The dataset is divided into four distinct clips, namely cordova1 (250 frames), cordova2 (406 frames), washington1 (337 frames), and washington2 (232 frames). For our purposes, we are utilizing the first three sets as the training set, while reserving the last set for testing. The dataset also provides 5 lane types: solid-white, solid-yellow, dashed-white, dashed-yellow, and double-yellow.

Our LVLane dataset matches our operating domain and provides images in urban scenarios with lane location and the same 7 lane types as used in TuSimple.  Images are 1200x720 with 464 training, 200 validation, and 294 testing images.

\begin{table}[b]
\caption{Datasets Summary}
\label{tab: datasets}
\begin{center}
\begin{tabular}{|c|c|c|c|c|c|} 
 \hline
 \textbf{Name} & \textbf{\#lanes} & \textbf{\#train} & \textbf{\#validation} & \textbf{\#test} & \textbf{Resolution}\\ 
 \hline
 TuSimple & $\leq 5$ & 3268 & 358 &2782 & 1280x720\\ 
 Caltech & $\leq 5$ & 993 & - & 232 & 640x480\\ 
 LVLane & $\leq 6$ & 464 & 200 & 294 & 1280x720\\ 
 \hline
\end{tabular}
\end{center}
\end{table}

\subsection{Evaluation Metrics}
For lane detection, we use the main evaluation metric of the TuSimple dataset, accuracy. It is defined as $$ACC_{lane} = \frac{\sum_{clip} C_{clip}}{\sum_{clip} S_{clip}}$$ where $C_{clip}$ is the correctly predicted lane points and $S_{clip}$ is the total number of ground truth lane points in each image. The result of all images is averaged to get the final accuracy. 

For the classification, we use accuracy too: $$ACC_{class} = \frac{\sum_{clip} TP_{clip}}{\sum_{clip} Total_{clip}}$$ where $TP_{clip}$ is the number of correctly predicted lane types (true positives) and $Total_{clip}$ is the total number of ground truth lane types in each image. Since our model predicts a maximum of 6 lanes including background (no lanes), while calculating evaluation metrics, we only consider non-background lane predictions (true positives) and the total number of non-background lanes (ground truths).

\subsection{Implementation Details}
We train three models using TuSimple, Caltech, and TuSimple+LVLane. Subsequently, we evaluate the models on their respective datasets as well as on cross-datasets. Here, we consider double-dashed as the dashed type of markings as they are mostly similar. So, we map 7 classes into 6 during training and testing. We focus mainly on training and testing the model for 2 classes since differentiating between solid and dashed is most crucial. The solid class includes solid-white, solid-yellow, double-yellow, and road edges whereas the dashed class includes dashed, double-dashed, and Botts' dots. %Besides, we report some measures for 6 classes too. 

We use data augmentation techniques like random rotation, horizontal flip, scaling, and translation to increase the robustness of the training data. It also prevents overfitting and increases the generalization ability of the model. In the optimization process, we use Stochastic Gradient Descent with a momentum of 0.9, weight decay 1e-4, and initial learning of 0.025. Besides, we use a custom learning rate scheduler that lowers the learning rate after each epoch. We train for 40 epochs and find that sufficient. 

All models are trained in a Ubuntu 20.04 machine with Python 3.8 and PyTorch 1.9.0 \cite{paszke2019pytorch}. The machine is equipped with a Nvidia Quadro RTX 6000 GPU. The full training loss consists of detection loss and classification loss.
\begin{equation}
    L_{total} =  L_{detection} + \gamma*L_{classification}
\end{equation}
%We set the weight to 0.6 and find it optimal for both tasks' optimal performance and faster convergence.
We found $\gamma=0.6$ for optimized detection and classification performance. 

\begin{figure}[bt]
     \begin{subfigure}{0.24\textwidth}
         \includegraphics[width=\textwidth,height=\myheight cm]{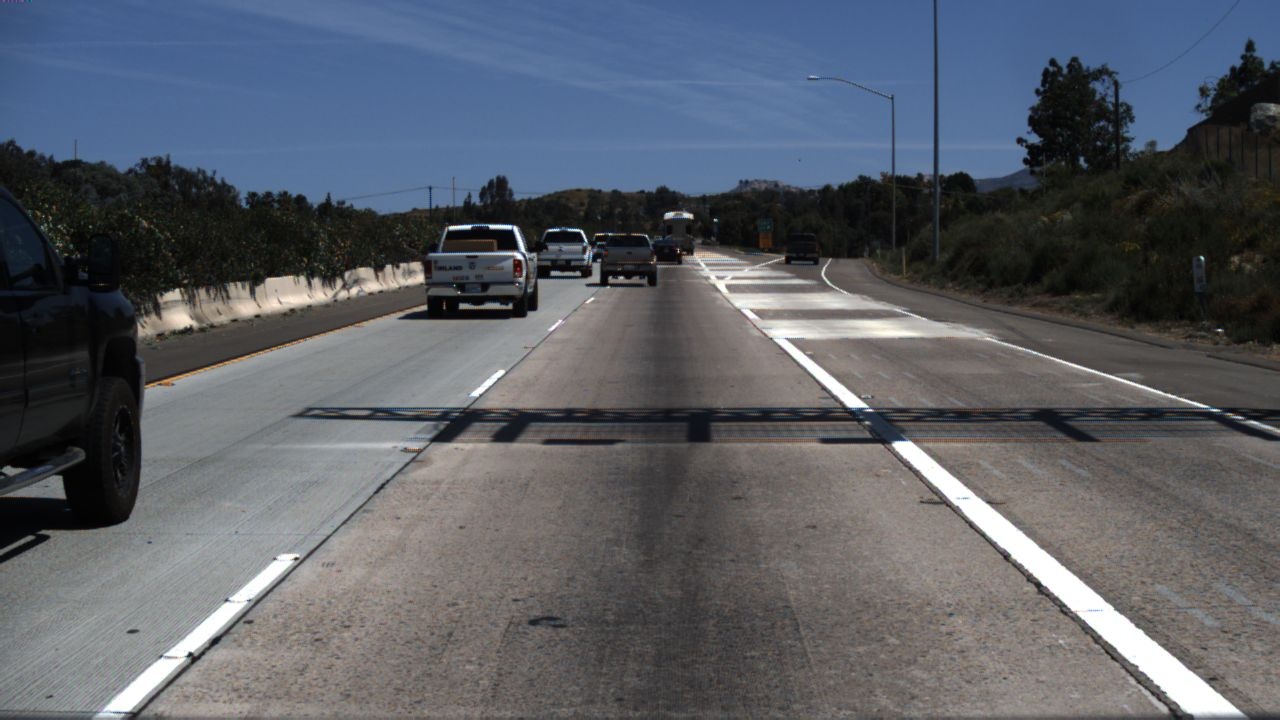}
     \end{subfigure}
     \begin{subfigure}{0.24\textwidth}
         \includegraphics[width=\textwidth, height=\myheight cm]{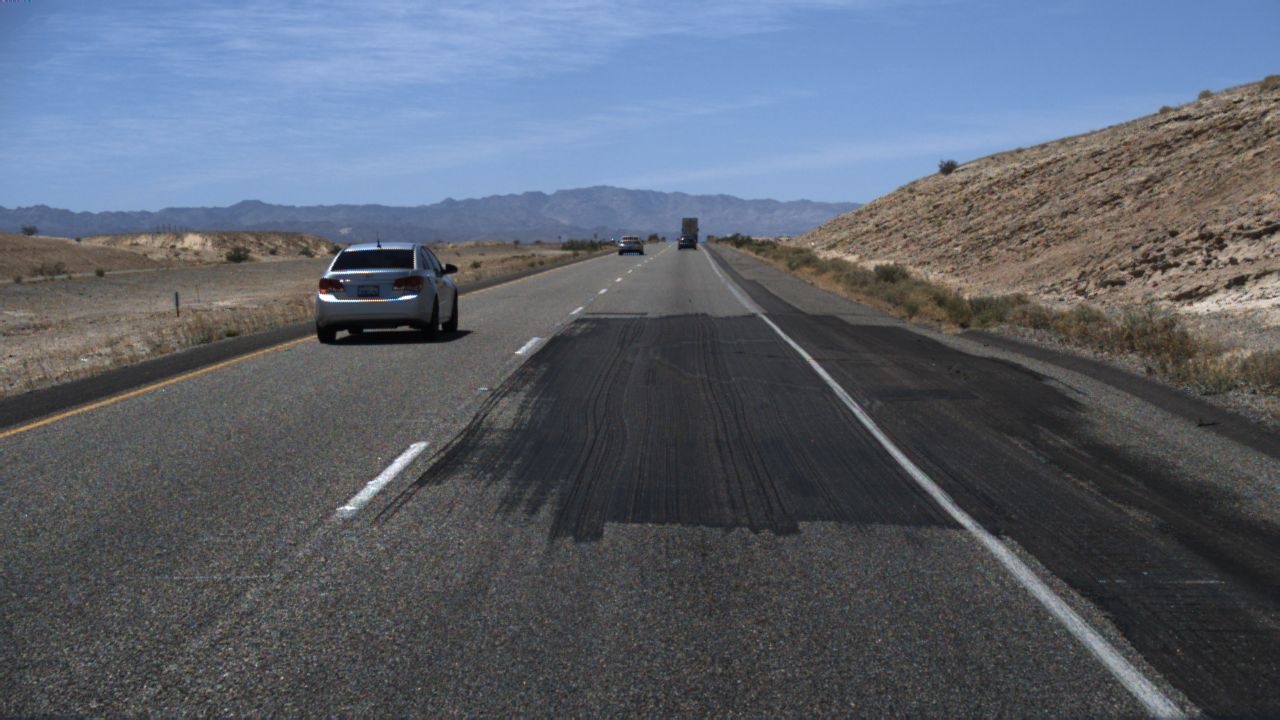}
     \end{subfigure}
     
    \vspace{0.2cm}
     \begin{subfigure}{0.24\textwidth}
         \includegraphics[width=\textwidth, height=\myheight cm]{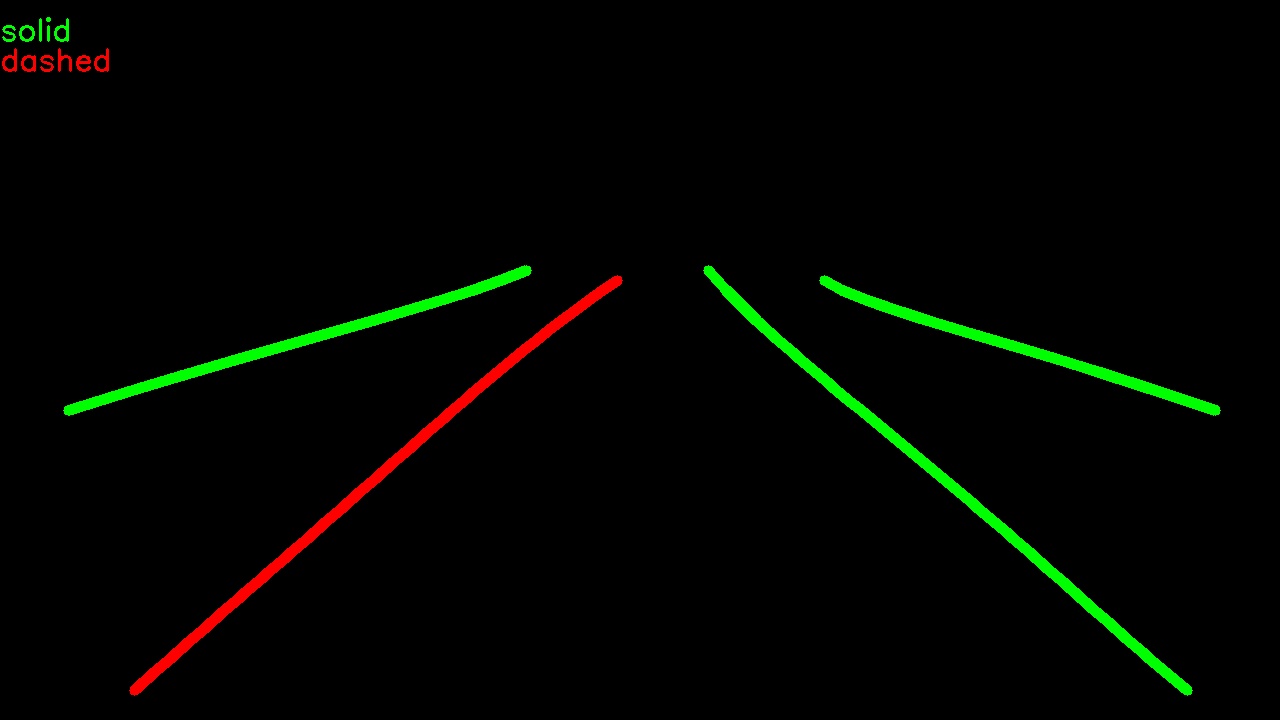}
     \end{subfigure}
     \begin{subfigure}{0.24\textwidth}
         \includegraphics[width=\textwidth, height=\myheight cm]{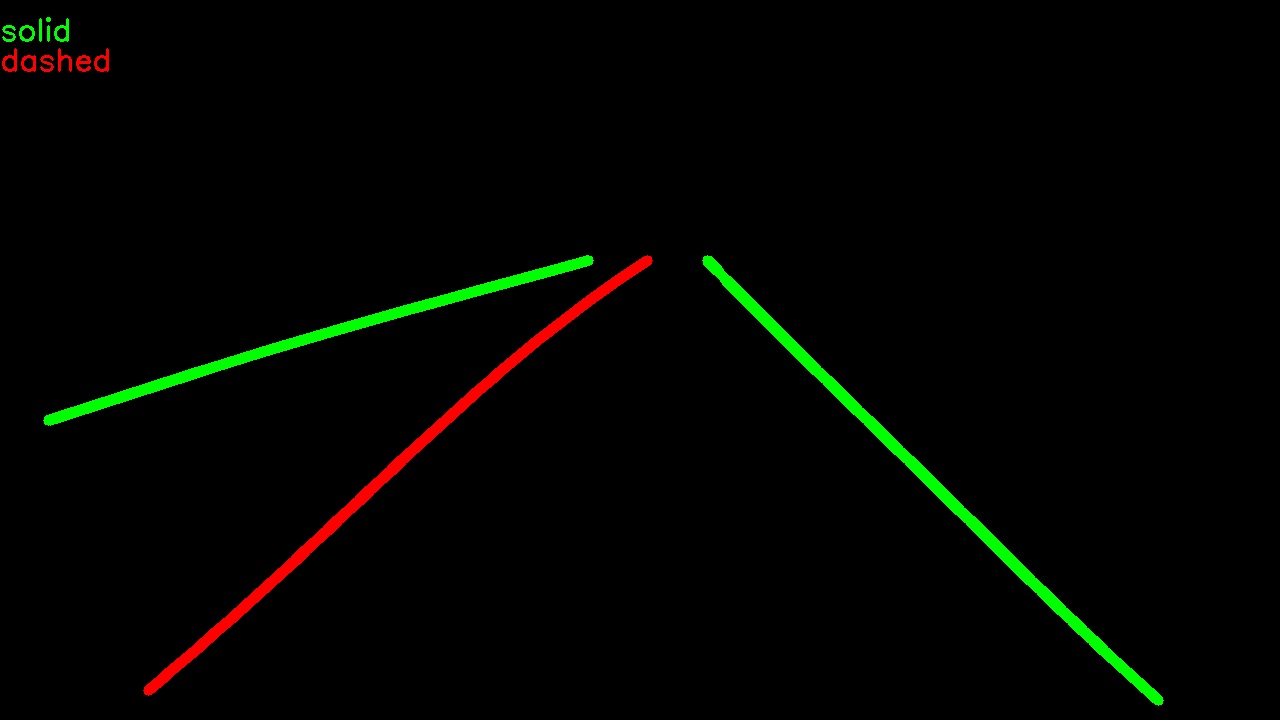}
     \end{subfigure}
     
     \vspace{0.2cm}
     \begin{subfigure}{0.24\textwidth}
         \includegraphics[width=\textwidth, height=\myheight cm]{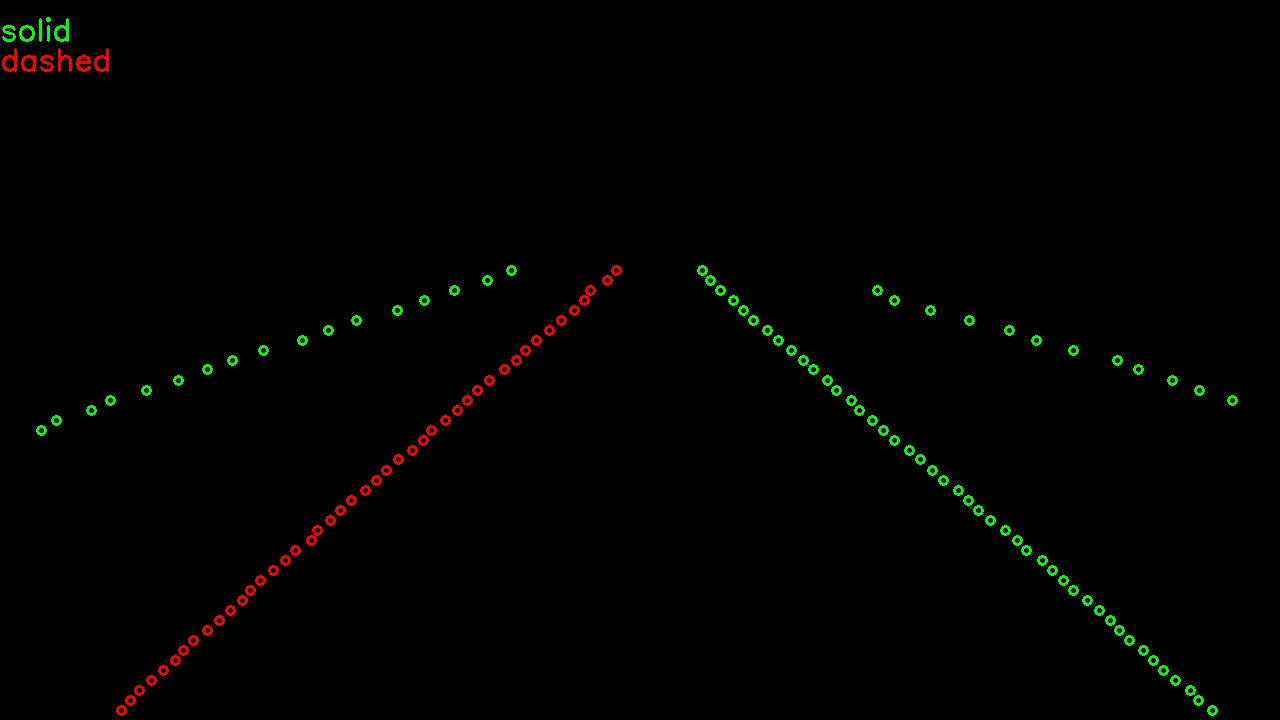}
     \end{subfigure}
     \begin{subfigure}{0.24\textwidth}
         \includegraphics[width=\textwidth, height=\myheight cm]{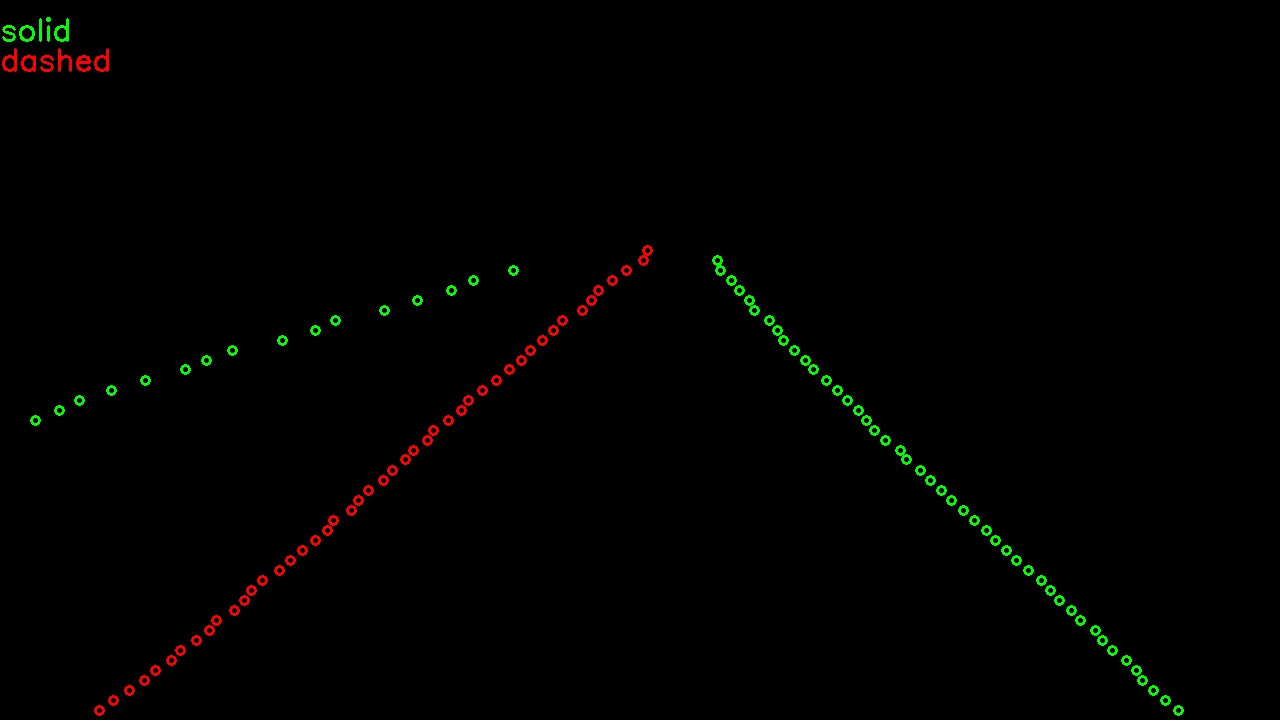}
     \end{subfigure}
        \caption{Classification performance on TuSimple test images. From top to bottom: original images, ground truth classes (solid - green, dashed - red), output images}
        \label{fig: test-class-tusimple-ufld}
\end{figure}

\subsection{Results}
\begin{figure}[bt]
     \begin{subfigure}{0.24\textwidth}
         \includegraphics[width=\textwidth,height=\myheight cm]{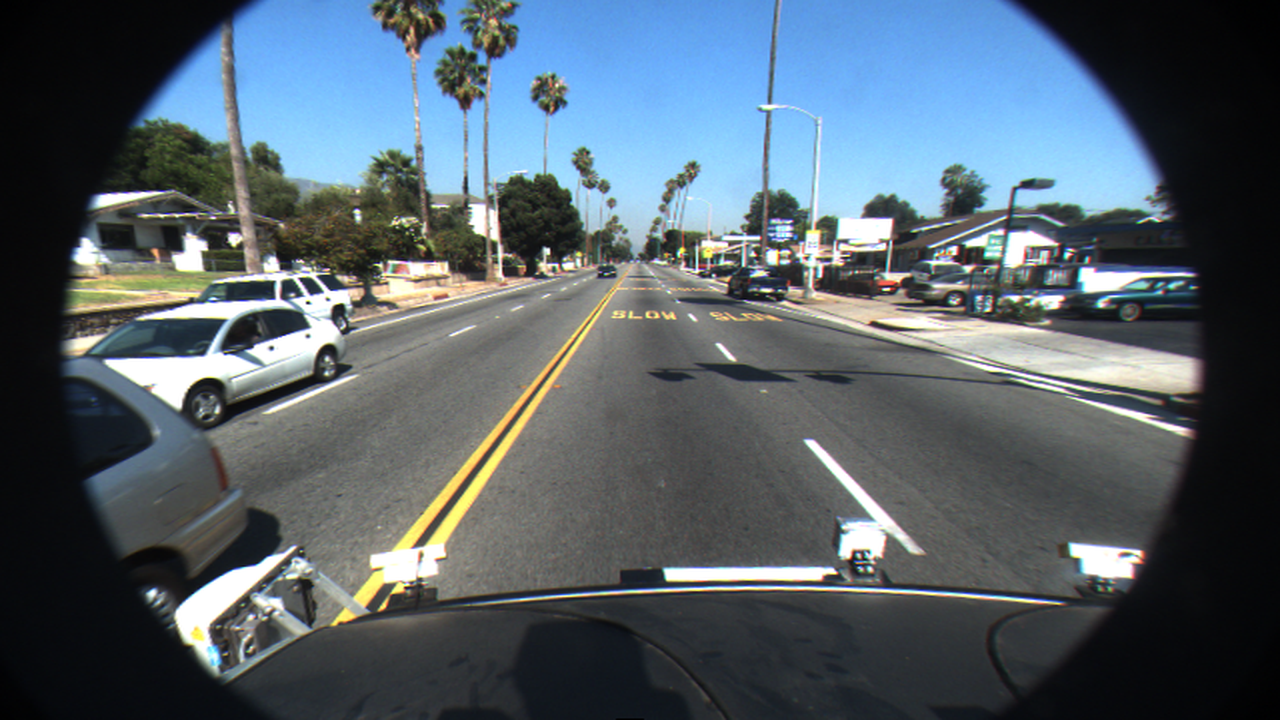}
     \end{subfigure}
     \begin{subfigure}{0.24\textwidth}
         \includegraphics[width=\textwidth, height=\myheight cm]{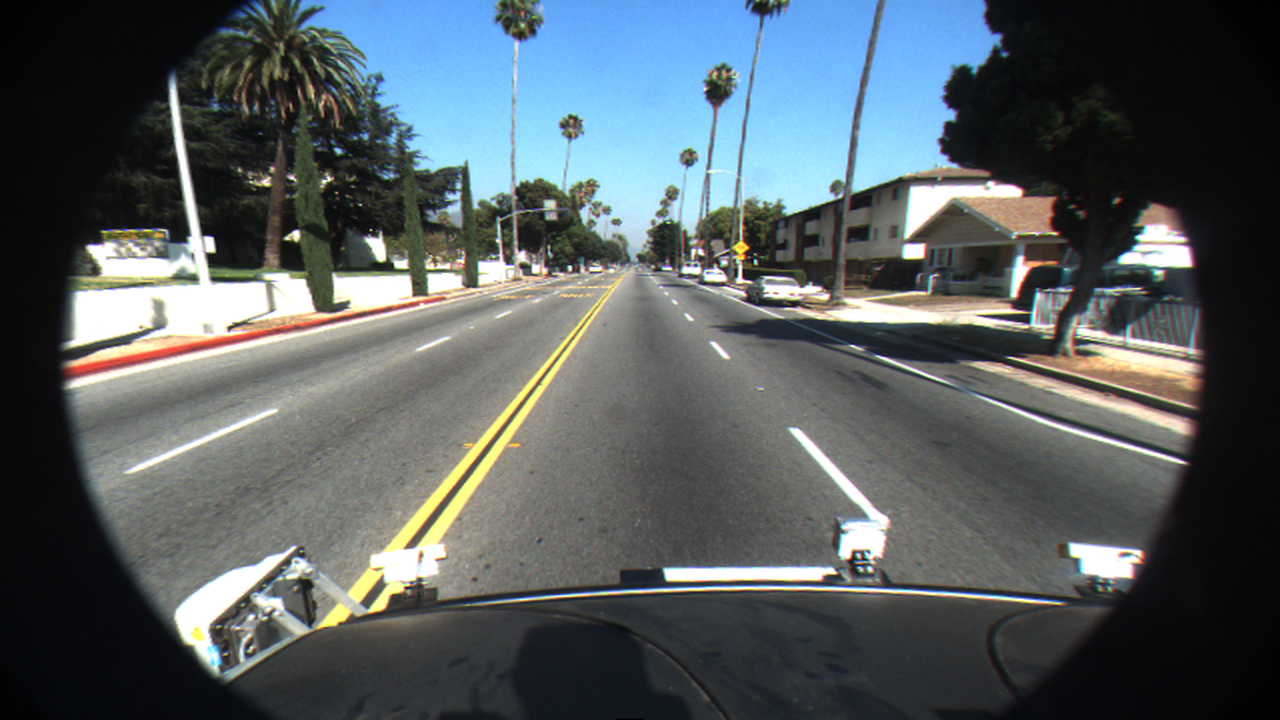}
     \end{subfigure}
     
    \vspace{0.2cm}
     \begin{subfigure}{0.24\textwidth}
         \includegraphics[width=\textwidth, height=\myheight cm]{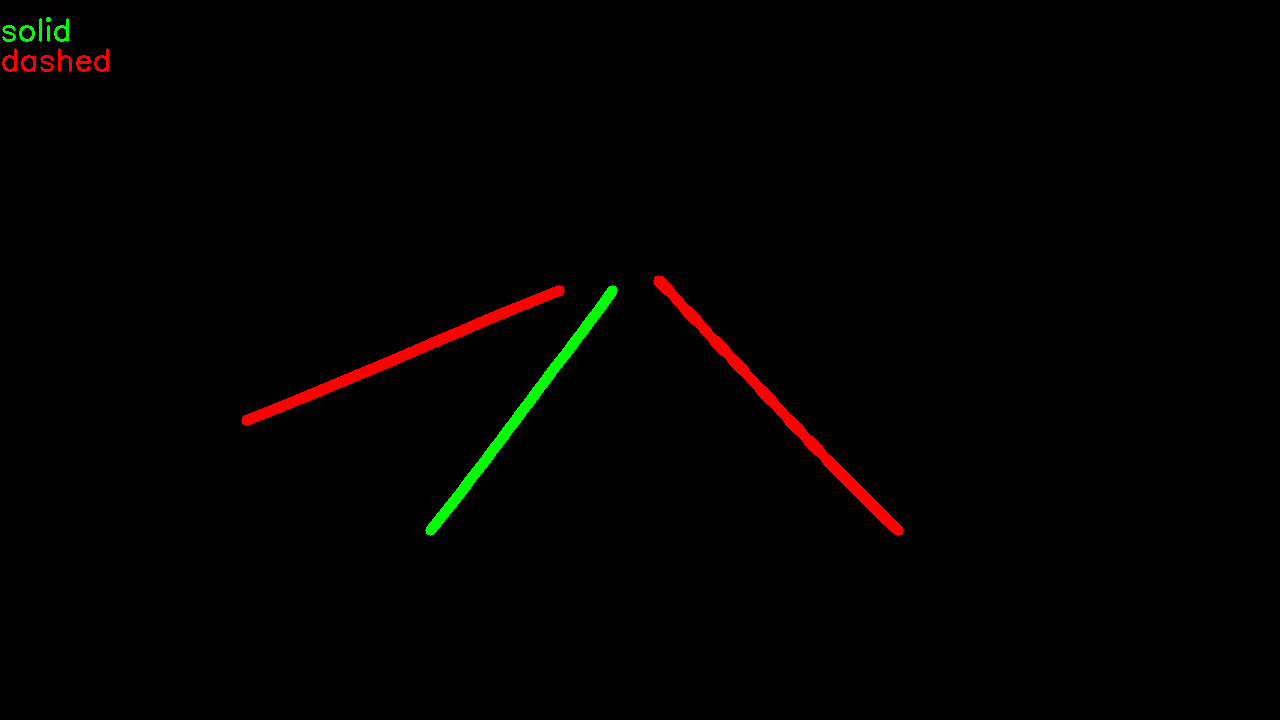}
     \end{subfigure}
     \begin{subfigure}{0.24\textwidth}
         \includegraphics[width=\textwidth, height=\myheight cm]{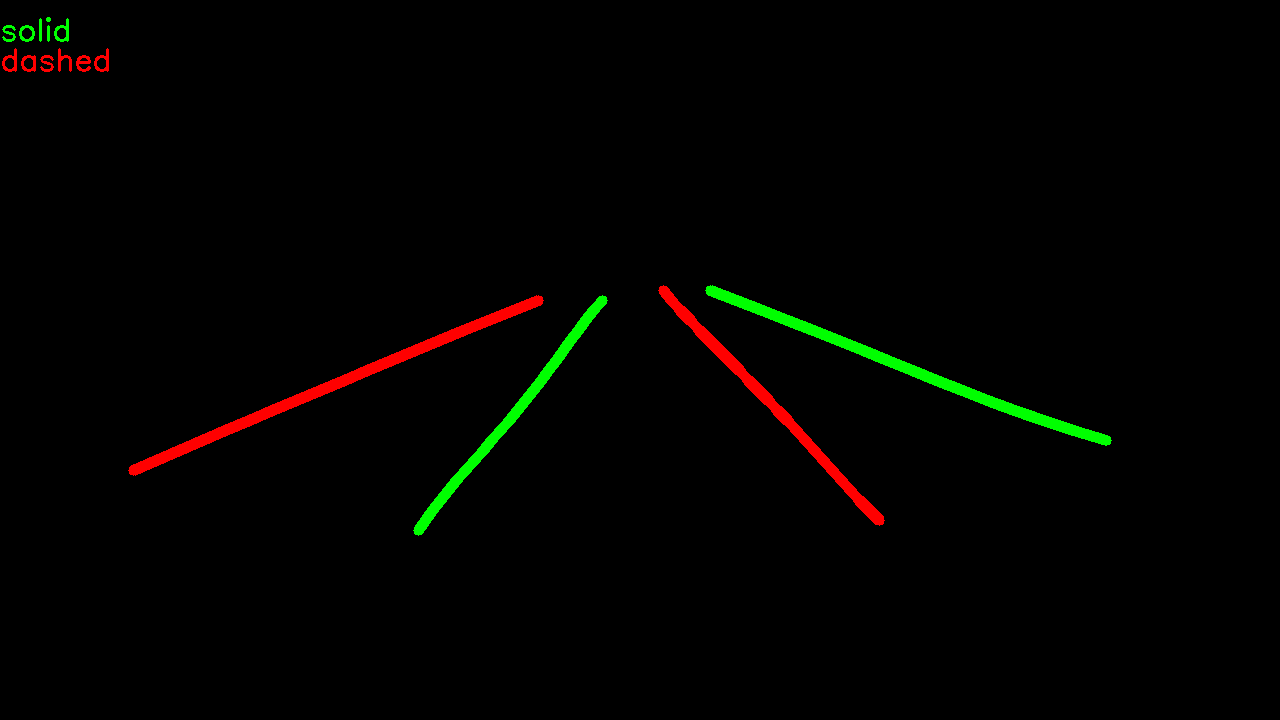}
     \end{subfigure}
     
     \vspace{0.2cm}
     \begin{subfigure}{0.24\textwidth}
         \includegraphics[width=\textwidth, height=\myheight cm]{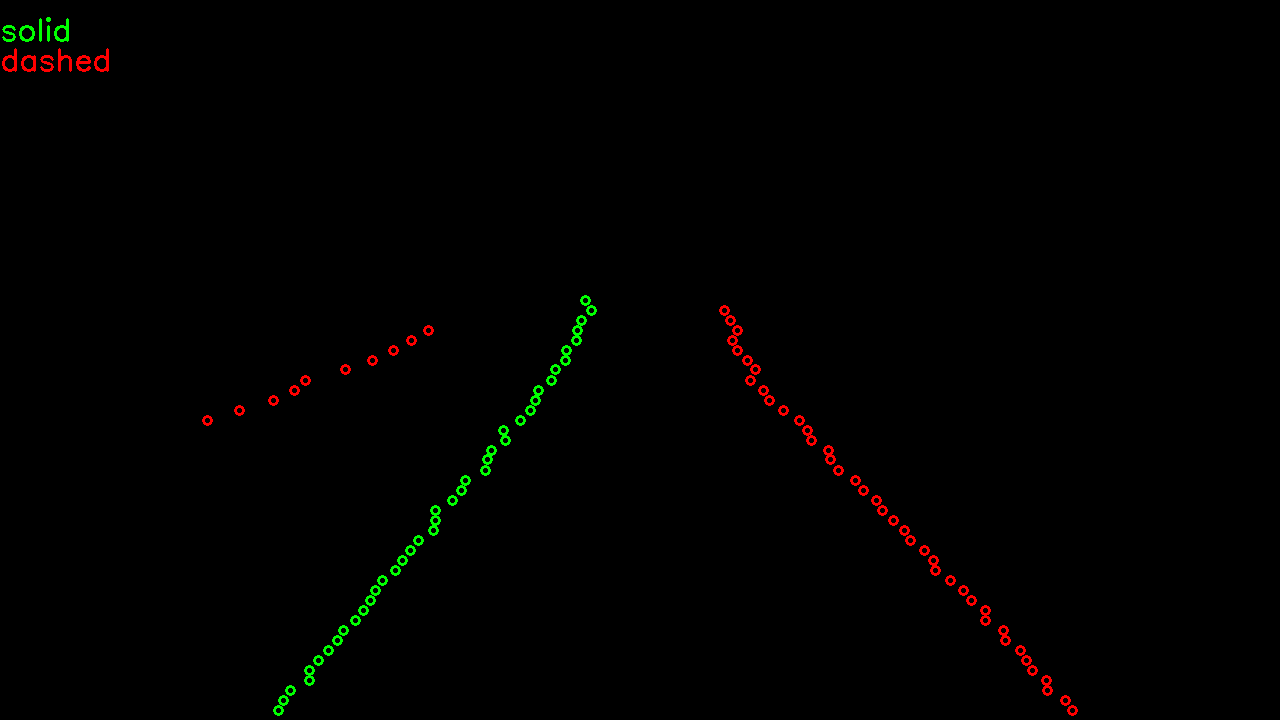}
     \end{subfigure}
     \begin{subfigure}{0.24\textwidth}
         \includegraphics[width=\textwidth, height=\myheight cm]{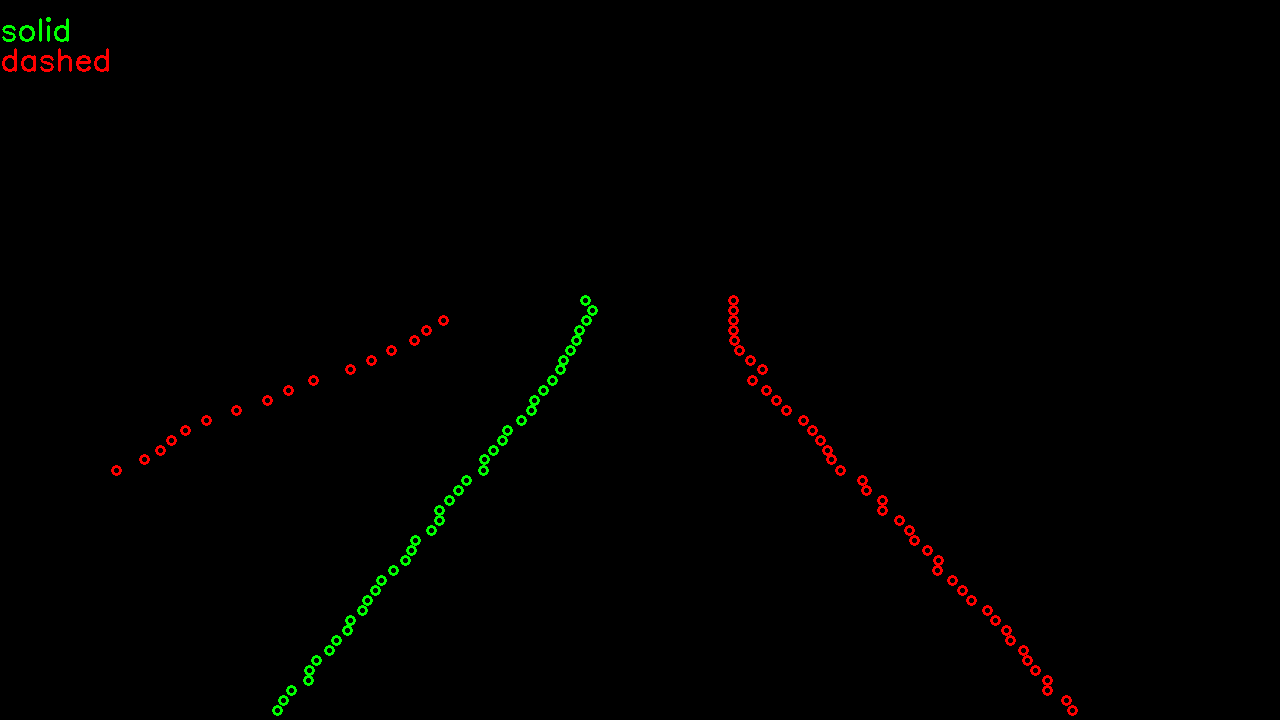}
     \end{subfigure}
        \caption{Classification performance on Caltech test images. From top to bottom: original images, ground truth classes (solid - green, dashed - red), output images}
        \label{fig: test-class-caltech-ufld}
\end{figure}

The findings of our study are summarized in Table \ref{tab: result}, which highlights the notable improvement in detection and classification performance on the LVLane test set when our dataset is utilized in addition to TuSimple for model training. Additionally, we observe the performance degradation when we assess the LVLane test set using a TuSimple-only trained model. TuSimple+LVLane was the top performer in detection and classification accuracy for the TuSimple test set. We report f1-scores of the classification branch to show the effect of false positives and false negatives. We notice an improvement in f1-scores commensurate with accuracy. The effectiveness of our model is evident in the results obtained on the TuSimple test set Fig. \ref{fig: test-class-tusimple-ufld} and Caltech test set Fig. \ref{fig: test-class-caltech-ufld}, where it demonstrates accurate detection and classification of solid and dashed lane types. 

\begin{table*}
\caption{Results Summary}
\label{tab: result}
\begin{center}
%\begin{tabular}{ |p{1cm}|p|p|p|p|p|p|p|p{1cm}| } 
\begin{tabular}{ |c|c|c|c|c|c|c|c|c| } 
 \hline
 \textbf{Method} & \textbf{Train set}& \textbf{Test set} & \textbf{detection} & \textbf{2-class} & \textbf{6-class} & \textbf{f1-score} & \thead{\textbf{training}\\ \textbf{time (hours)}} & \thead{\textbf{inference}\\ \textbf{time (ms)}}\\ 
 \hline
 UFLD & TuSimple train & TuSimple test & 0.934 & 0.935 & 0.929 & 0.934 & 0.7 & 7.6\\ 
 UFLD & TuSimple train & LVLane test & 0.720 & 0.660 & - & 0.672 & 0.7 & 7.6\\ 
 UFLD (Ours) & TuSimple+LVLane train & TuSimple test & 0.945 & 0.949 & 0.943 & 0.955 & 0.8 & 7.5\\
 UFLD (Ours) & TuSimple+LVLane train & LVLane test & 0.773 & 0.790 & - & 0.776 & 0.8 & 7.5\\
 \hline
 UFLD & Caltech train  & Caltech test & 0.813 & 0.930 & - & - & 0.4 & 7\\
 \hline
 UFLD-MP\tablefootnote{MP -- mixed precision} & TuSimple train & TuSimple test & 0.935 & 0.935 & 0.928 & - & 0.71 & 8.68\\
 RESA & TuSimple train & TuSimple test & 0.939 & 0.937 & 0.932 & - & 1.83 & 29.56\\
 RESA-MP & TuSimple train & TuSimple test & 0.942 & 0.937 & 0.931 & - & 1.3 & 22.12\\
 \hline
\end{tabular}
\end{center}
\end{table*}

\begin{figure*}[t]
    %\setlength\tabcolsep{0pt}
    %\begin{tabular*}{0.95\linewidth}%{@{\extracolsep{\fill}}ccccc}
    %Surface Reflection & Botts' Dots & Botts' Dots & Dense Traffic & No Lane Markers
    %\end{tabular*}\\
    \begin{subfigure}[b]{0.19\textwidth}
         \begin{center} Surface Reflection \end{center}
     \end{subfigure}
     \begin{subfigure}[b]{0.20\textwidth}
         \begin{center} Botts' Dots \end{center}
     \end{subfigure}
     \begin{subfigure}[b]{0.20\textwidth}
         \begin{center} Botts' Dots \end{center}
     \end{subfigure}
     \begin{subfigure}[b]{0.20\textwidth}
         \begin{center} Dense Traffic \end{center}
     \end{subfigure}
     \begin{subfigure}[b]{0.19\textwidth}
         \begin{center} No Lane Markers \end{center}
     \end{subfigure} \\[0.5em]
     \begin{subfigure}[b]{0.19\textwidth}
         \includegraphics[width=\textwidth, height=\myheight cm]{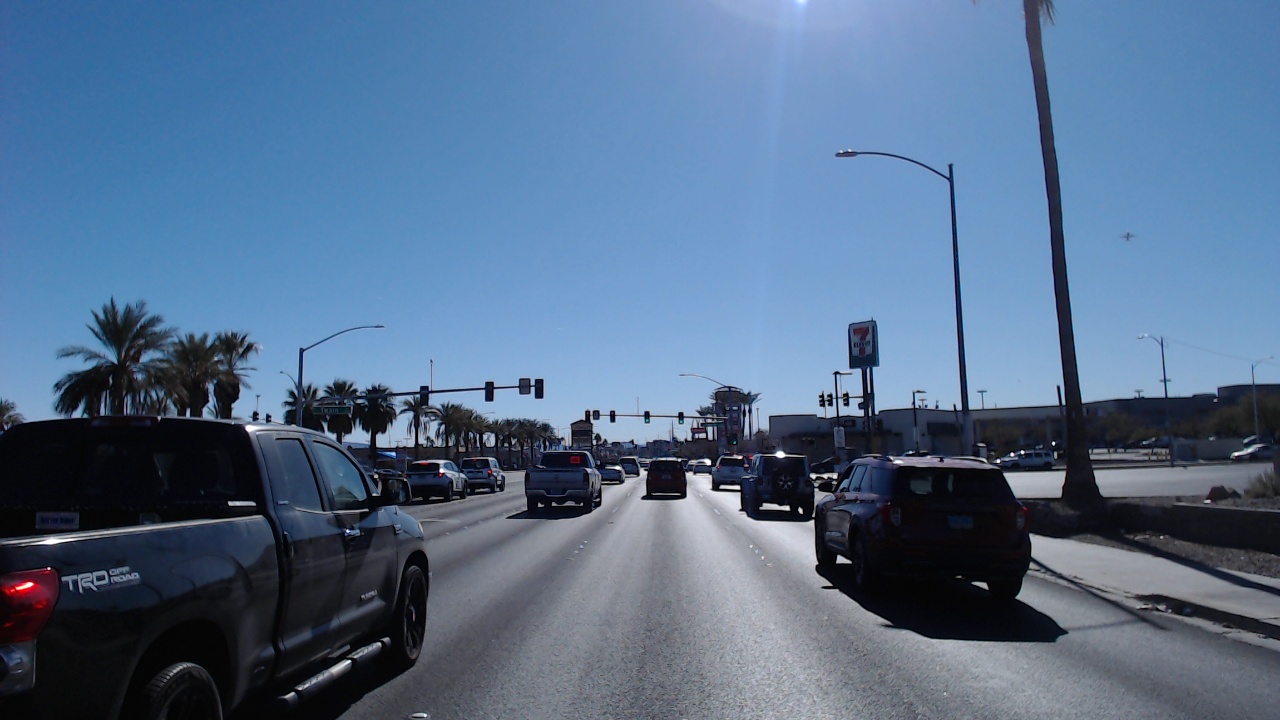}
     \end{subfigure}
     \begin{subfigure}[b]{0.20\textwidth}
         \includegraphics[width=\textwidth, height=\myheight cm]{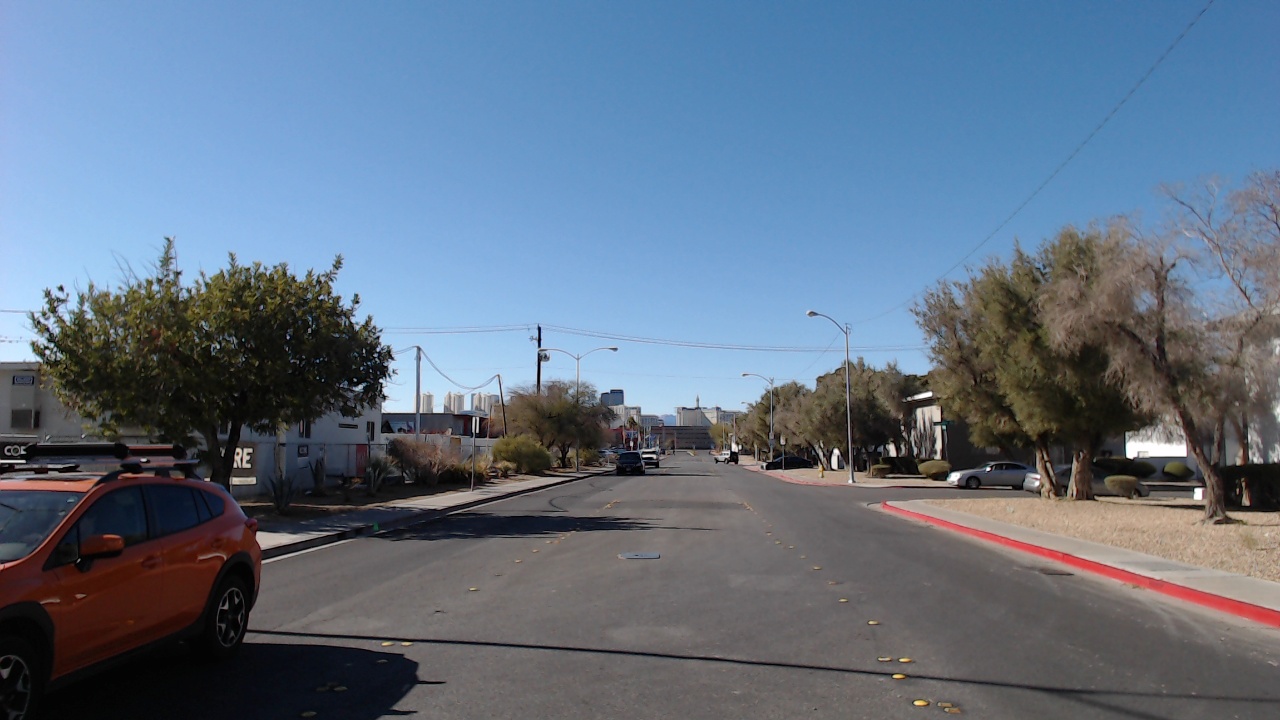}
     \end{subfigure}
     \begin{subfigure}[b]{0.20\textwidth}
         \includegraphics[width=\textwidth, height=\myheight cm]{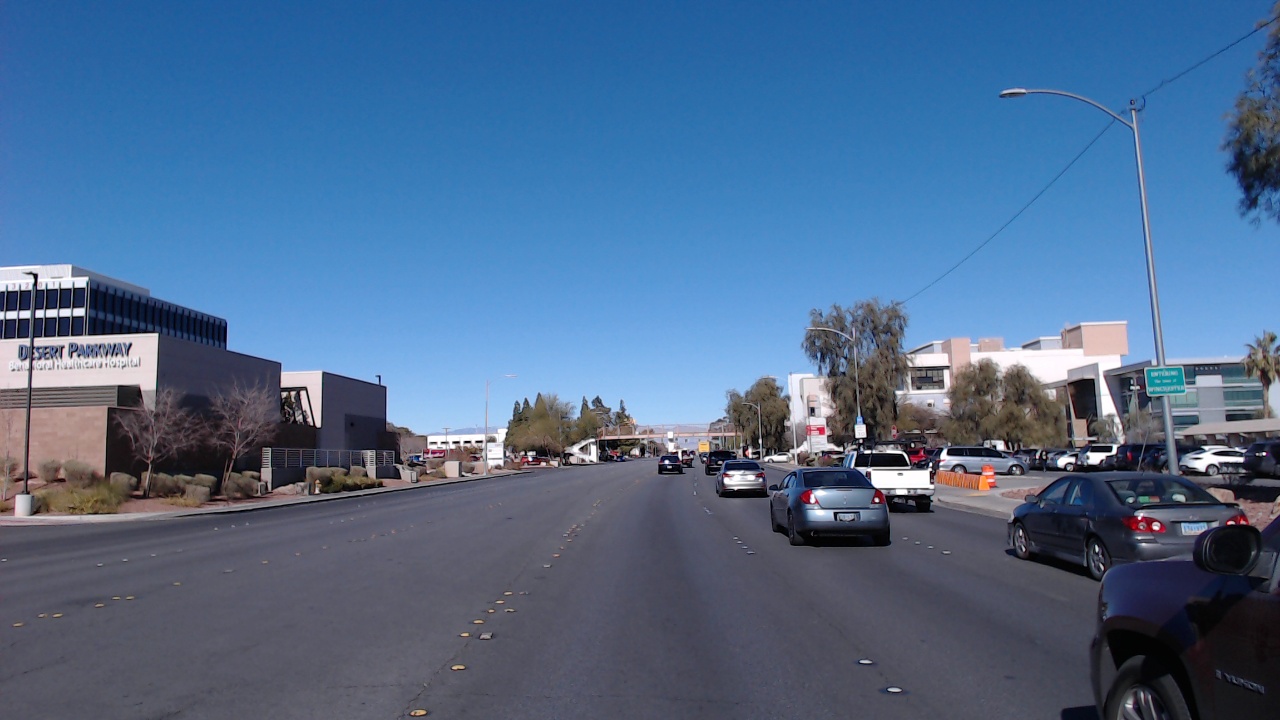}
     \end{subfigure}
     \begin{subfigure}[b]{0.20\textwidth}
         \includegraphics[width=\textwidth, height=\myheight cm]{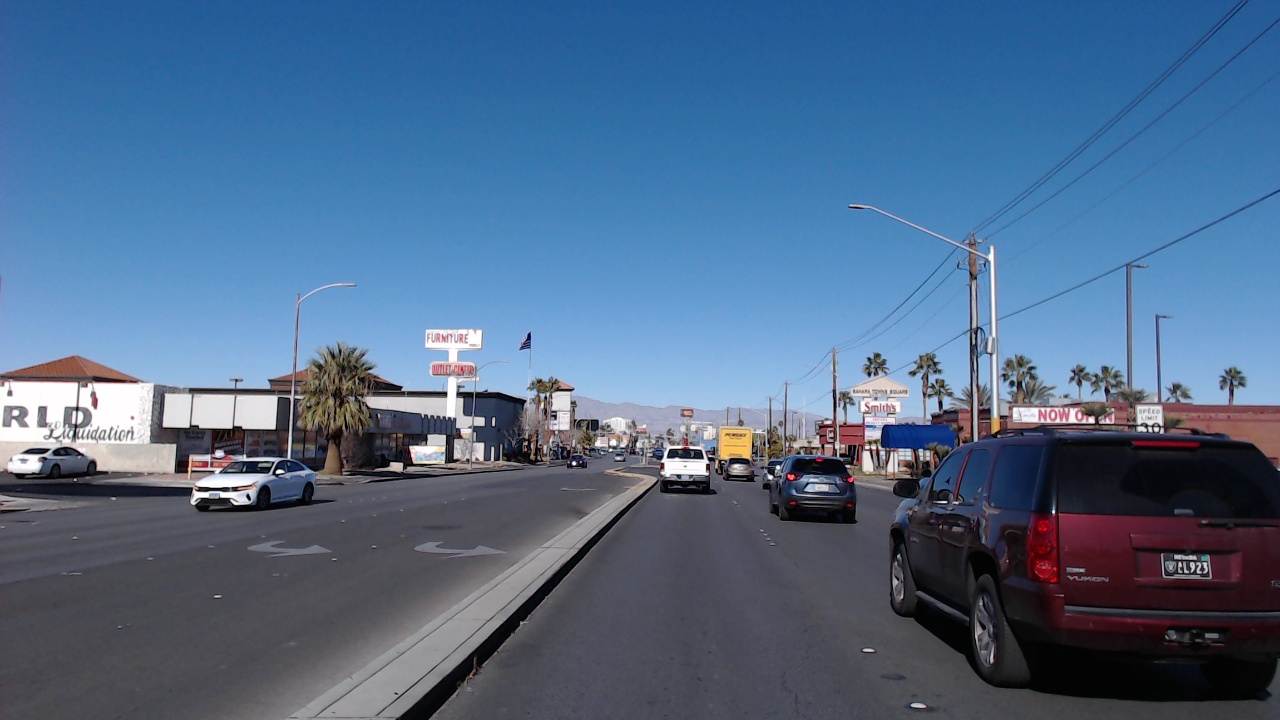}
     \end{subfigure}
     \begin{subfigure}[b]{0.19\textwidth}
         \includegraphics[width=\textwidth, height=\myheight cm]{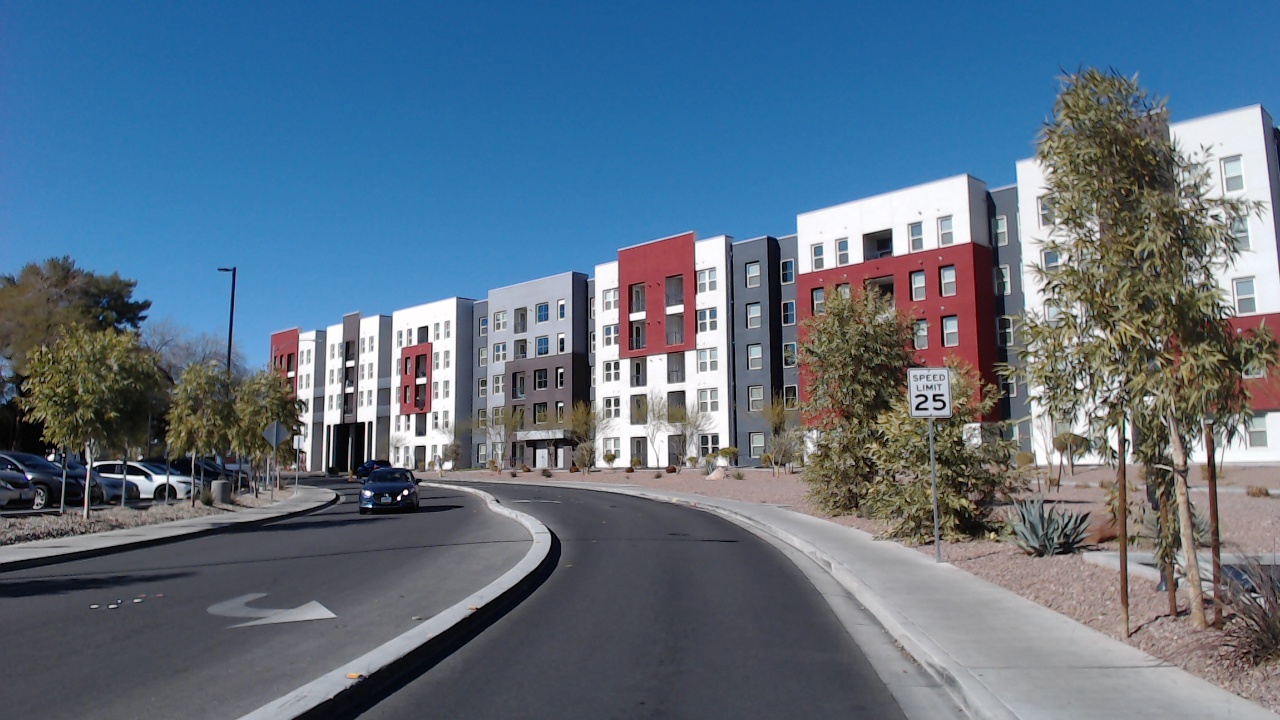}
     \end{subfigure}
     
     \vspace{0.2cm}
     \begin{subfigure}[b]{0.19\textwidth}
         \includegraphics[width=\textwidth, height=\myheight cm]{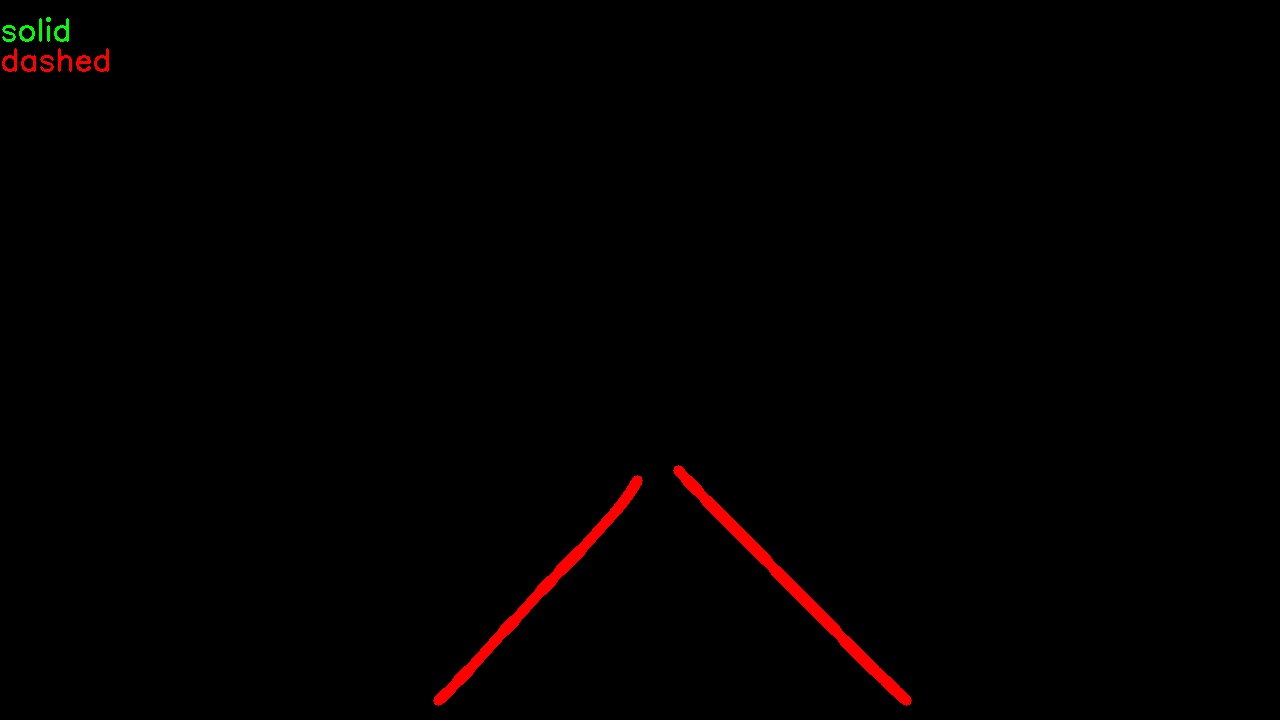}
     \end{subfigure}
     \begin{subfigure}[b]{0.20\textwidth}
         \includegraphics[width=\textwidth, height=\myheight cm]{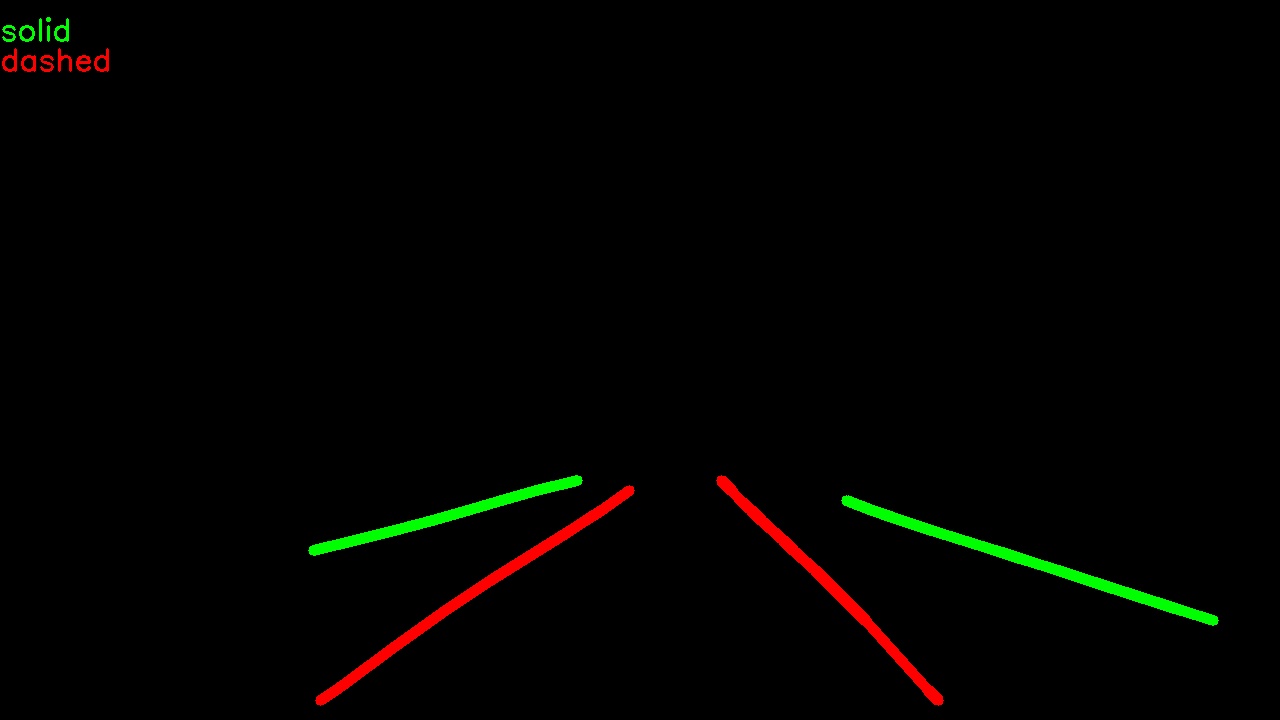}
     \end{subfigure}
     \begin{subfigure}[b]{0.20\textwidth}
         \includegraphics[width=\textwidth, height=\myheight cm]{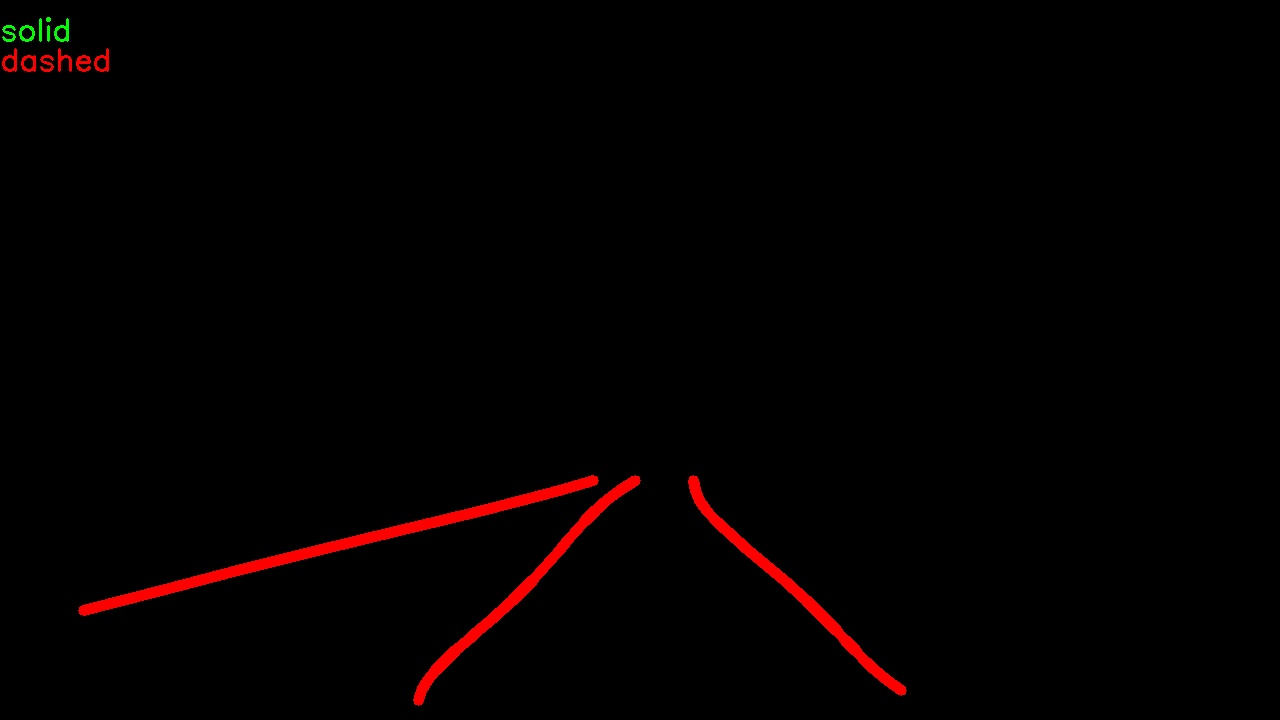}
     \end{subfigure}
     \begin{subfigure}[b]{0.20\textwidth}
         \includegraphics[width=\textwidth, height=\myheight cm]{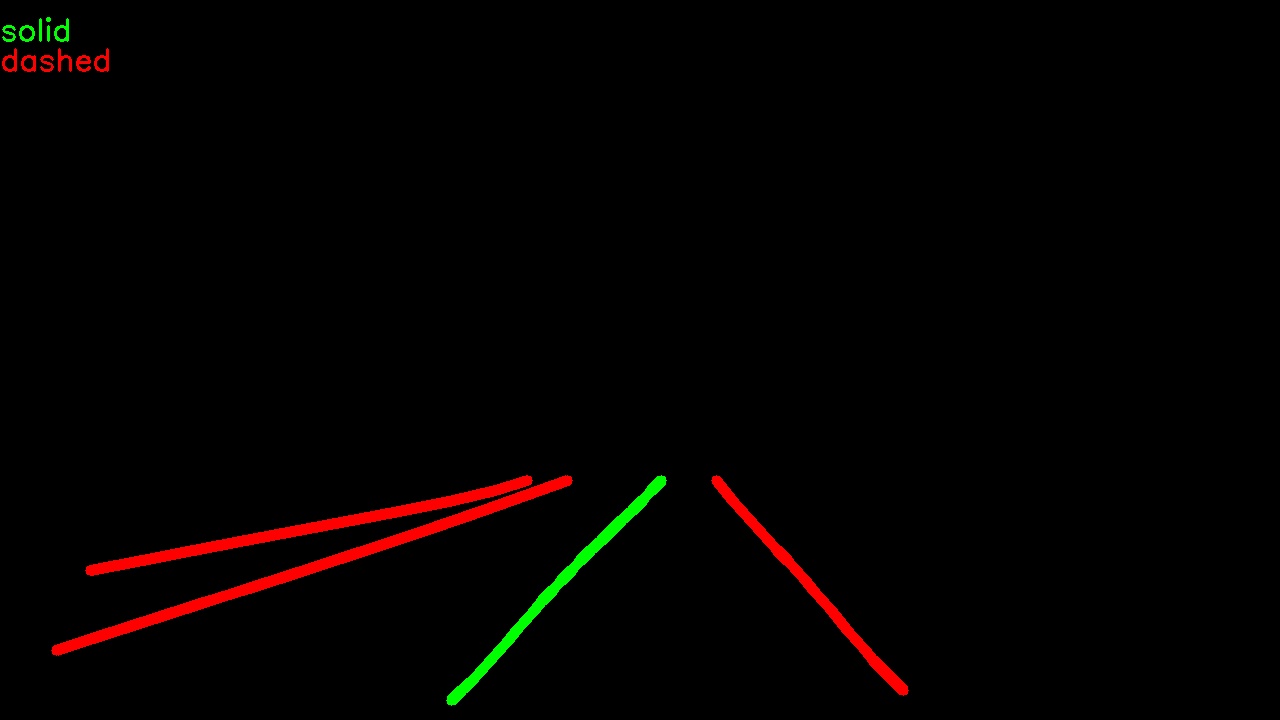}
     \end{subfigure}
     \begin{subfigure}[b]{0.19\textwidth}
         \includegraphics[width=\textwidth, height=\myheight cm]{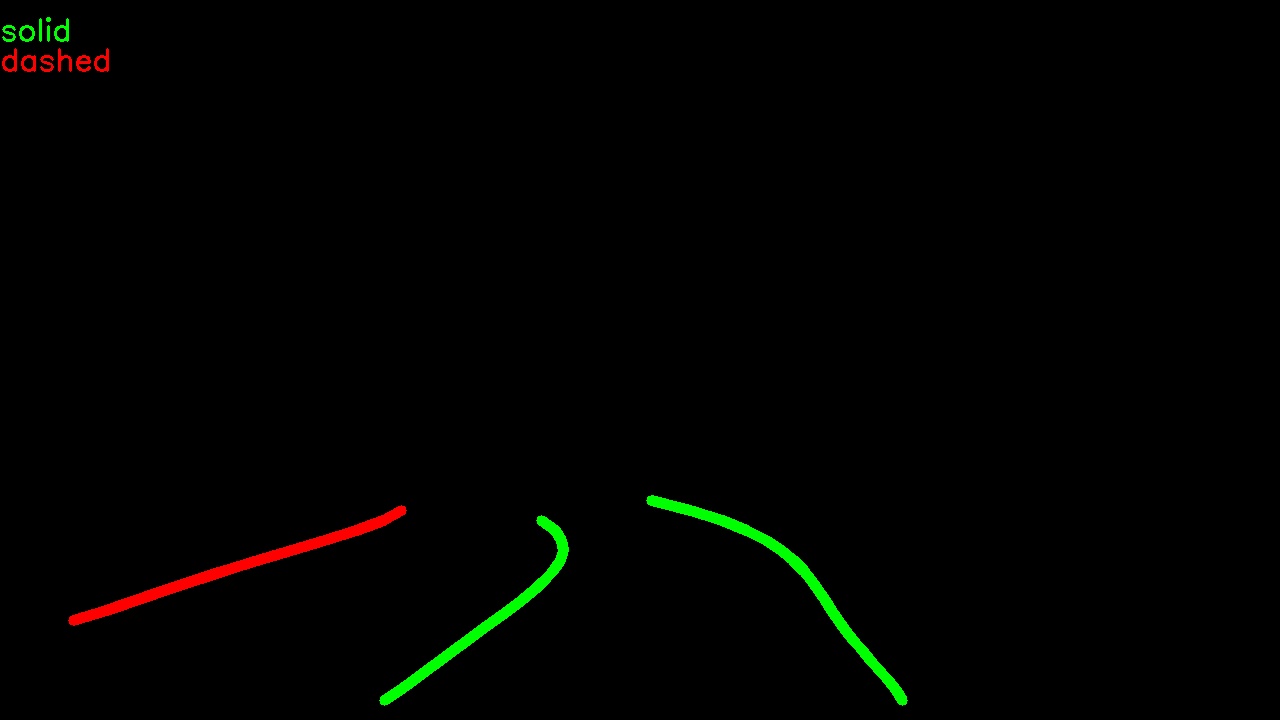}
     \end{subfigure}
     
     \vspace{0.2cm}
     \begin{subfigure}[b]{0.19\textwidth}
         \includegraphics[width=\textwidth, height=\myheight cm]{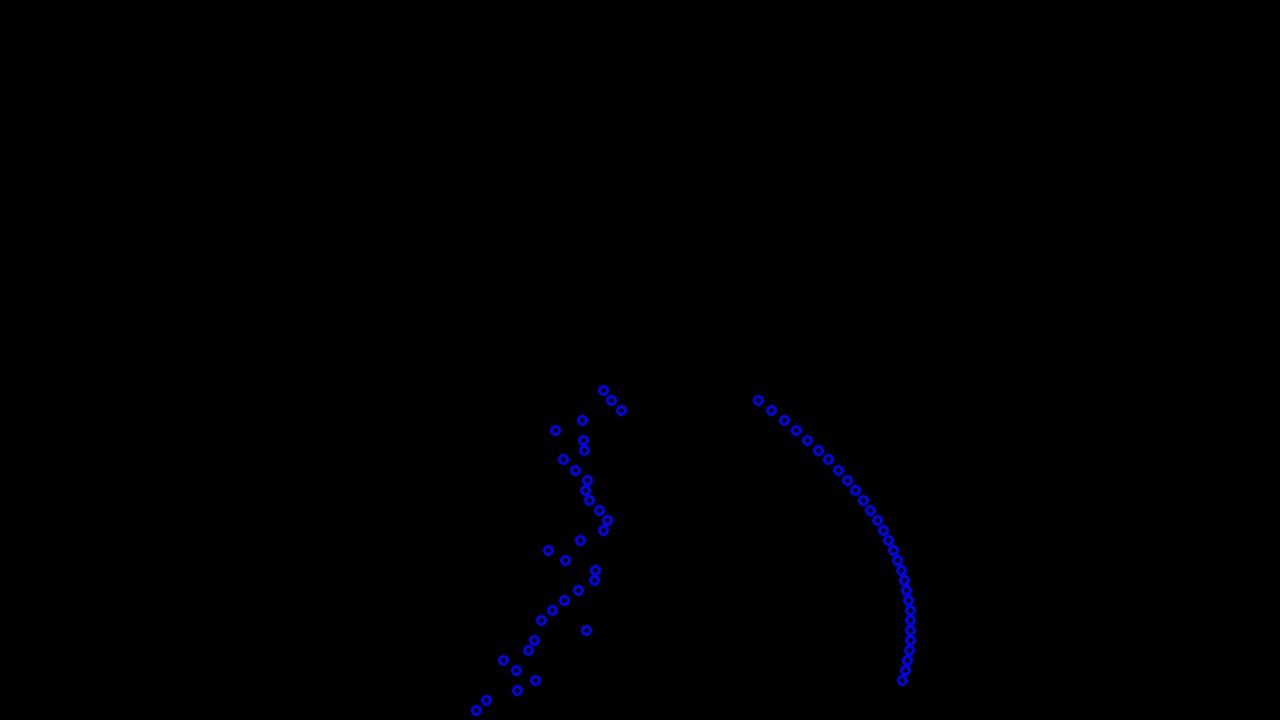}
     \end{subfigure}
     \begin{subfigure}[b]{0.20\textwidth}
         \includegraphics[width=\textwidth, height=\myheight cm]{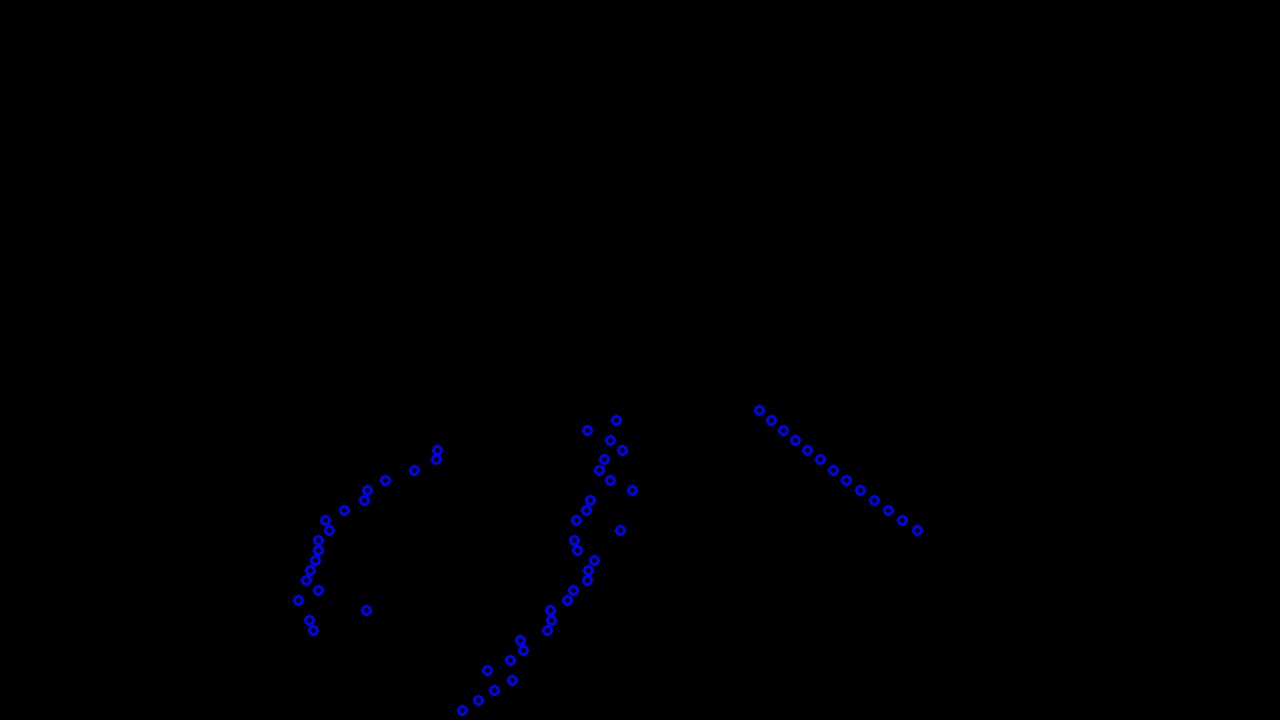}
     \end{subfigure}
     \begin{subfigure}[b]{0.20\textwidth}
         \includegraphics[width=\textwidth, height=\myheight cm]{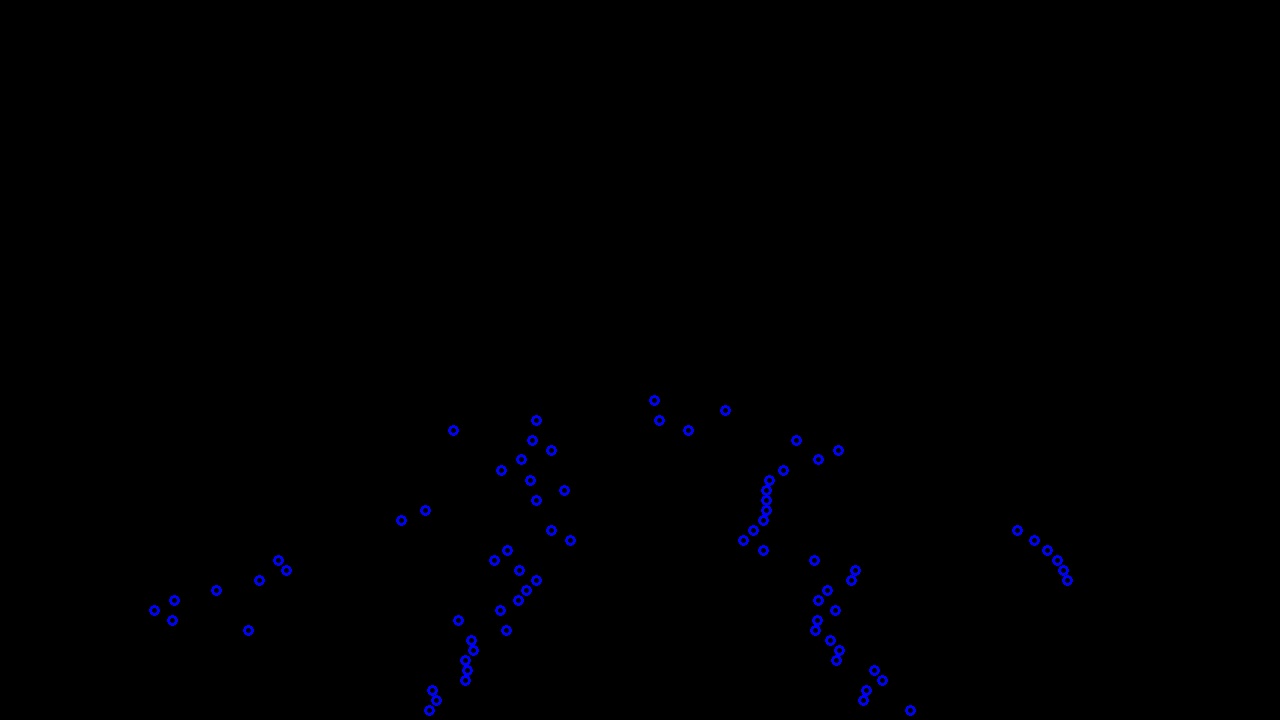}
     \end{subfigure}
     \begin{subfigure}[b]{0.20\textwidth}
         \includegraphics[width=\textwidth, height=\myheight cm]{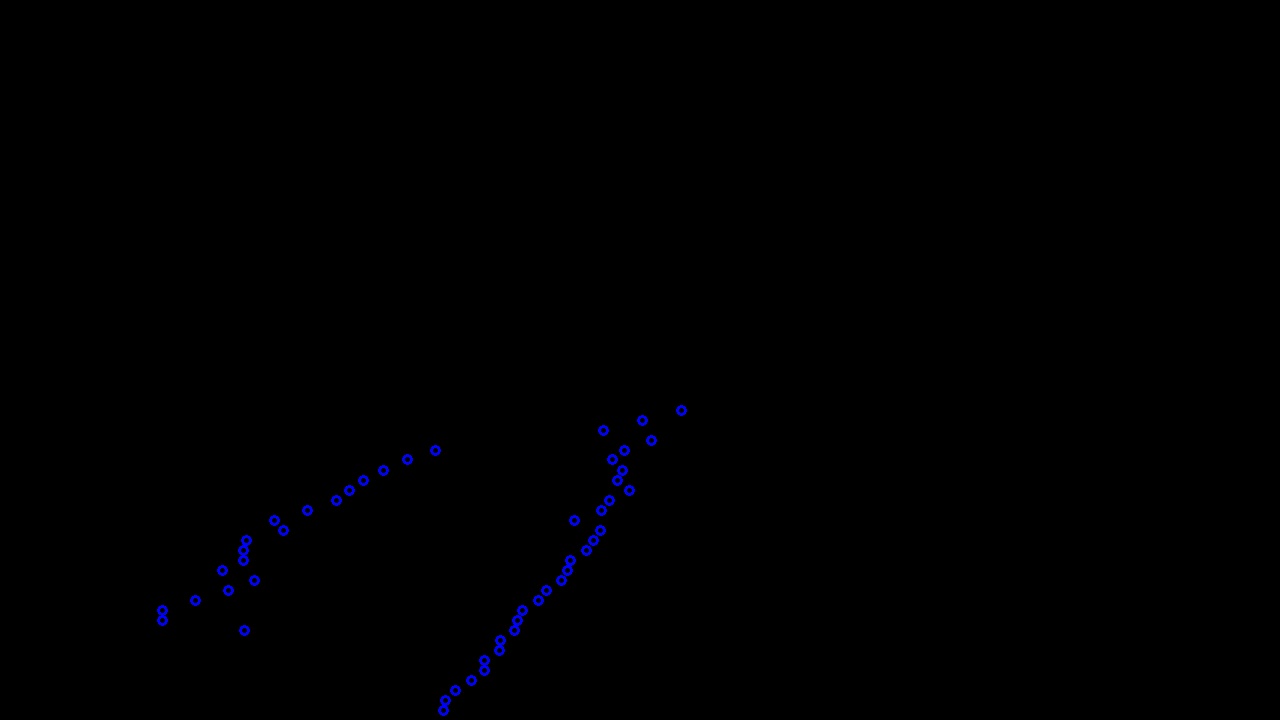}
     \end{subfigure}
     \begin{subfigure}[b]{0.19\textwidth}
         \includegraphics[width=\textwidth, height=\myheight cm]{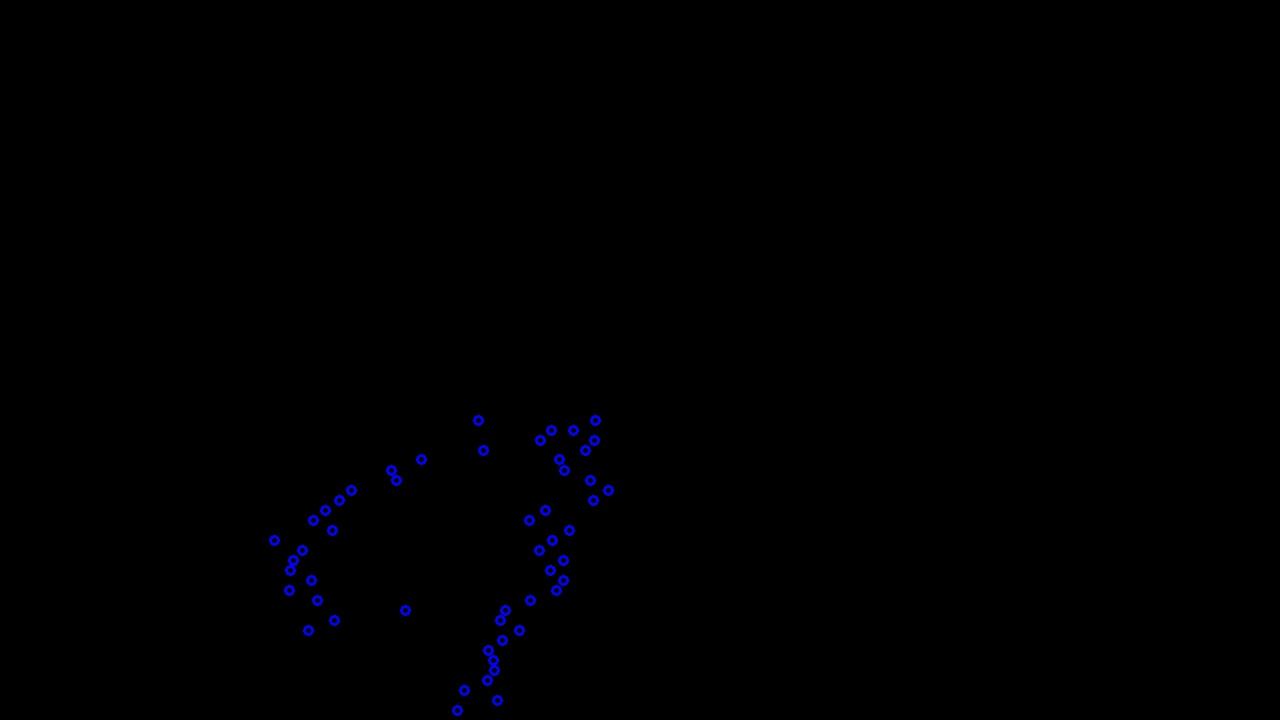}
     \end{subfigure}

     \vspace{0.2cm}
     \begin{subfigure}[b]{0.19\textwidth}
         \includegraphics[width=\textwidth, height=\myheight cm]{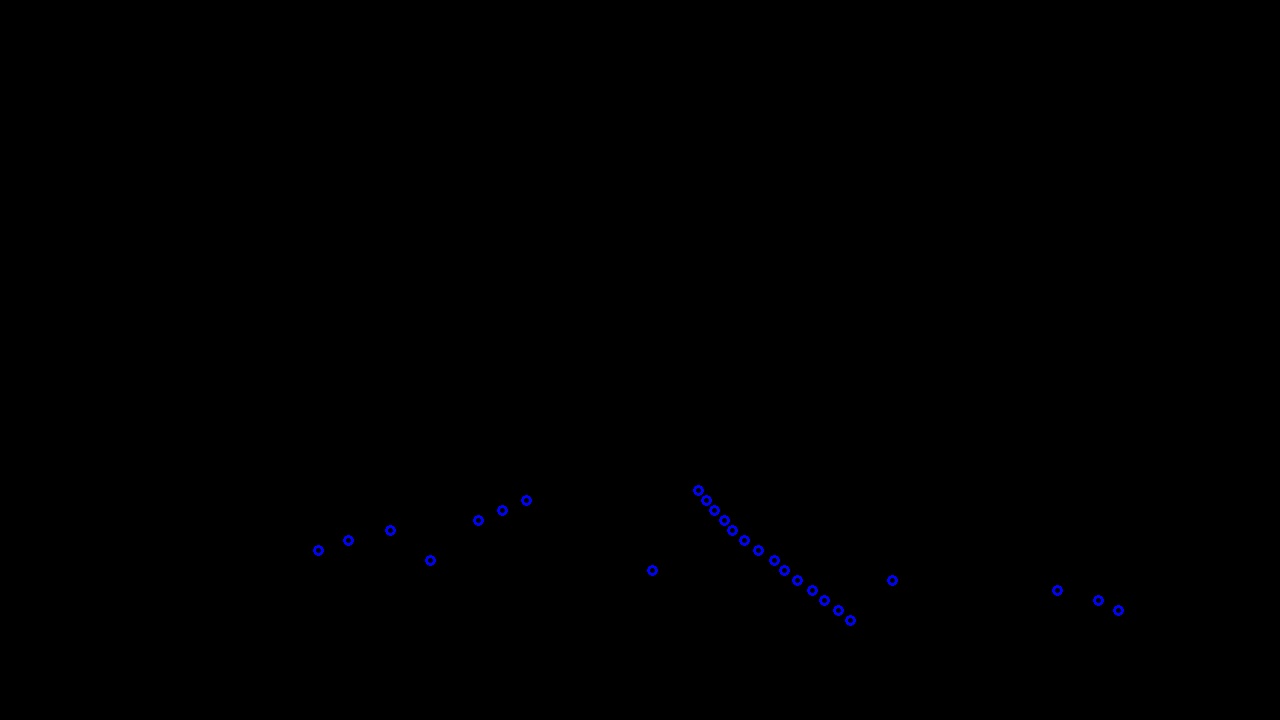}
     \end{subfigure}
     \begin{subfigure}[b]{0.20\textwidth}
         \includegraphics[width=\textwidth, height=\myheight cm]{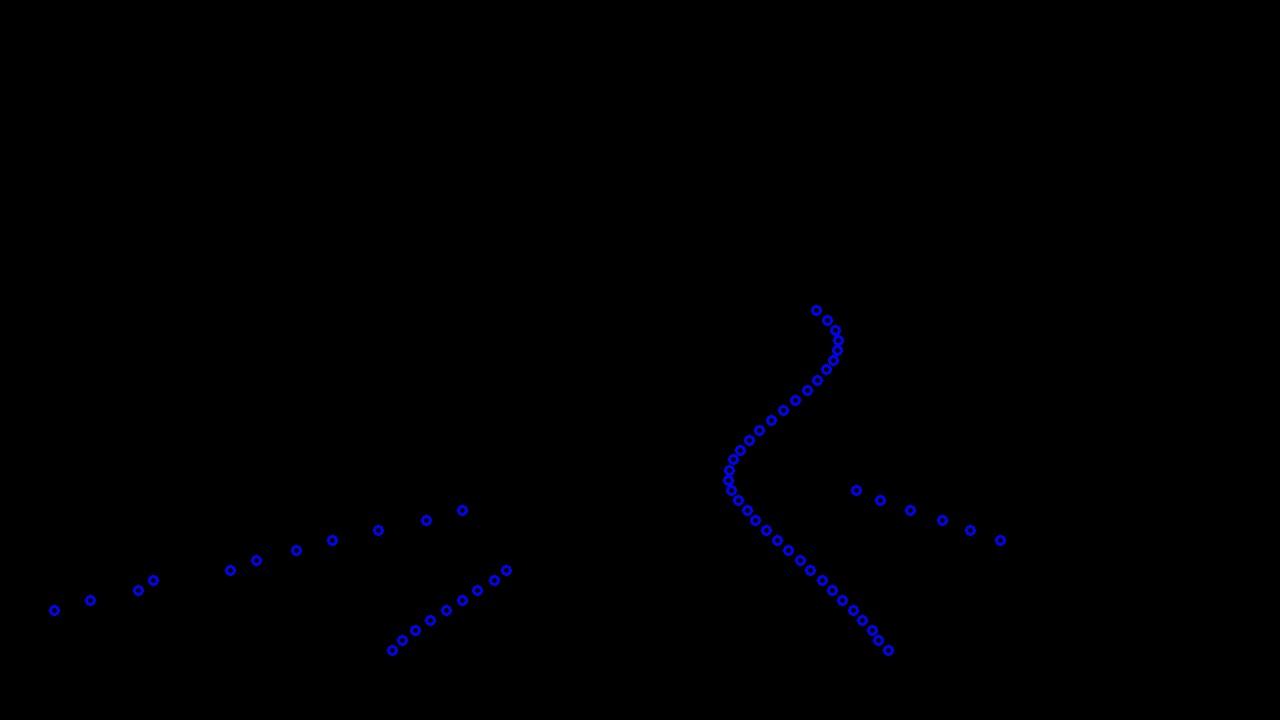}
     \end{subfigure}
     \begin{subfigure}[b]{0.20\textwidth}
         \includegraphics[width=\textwidth, height=\myheight cm]{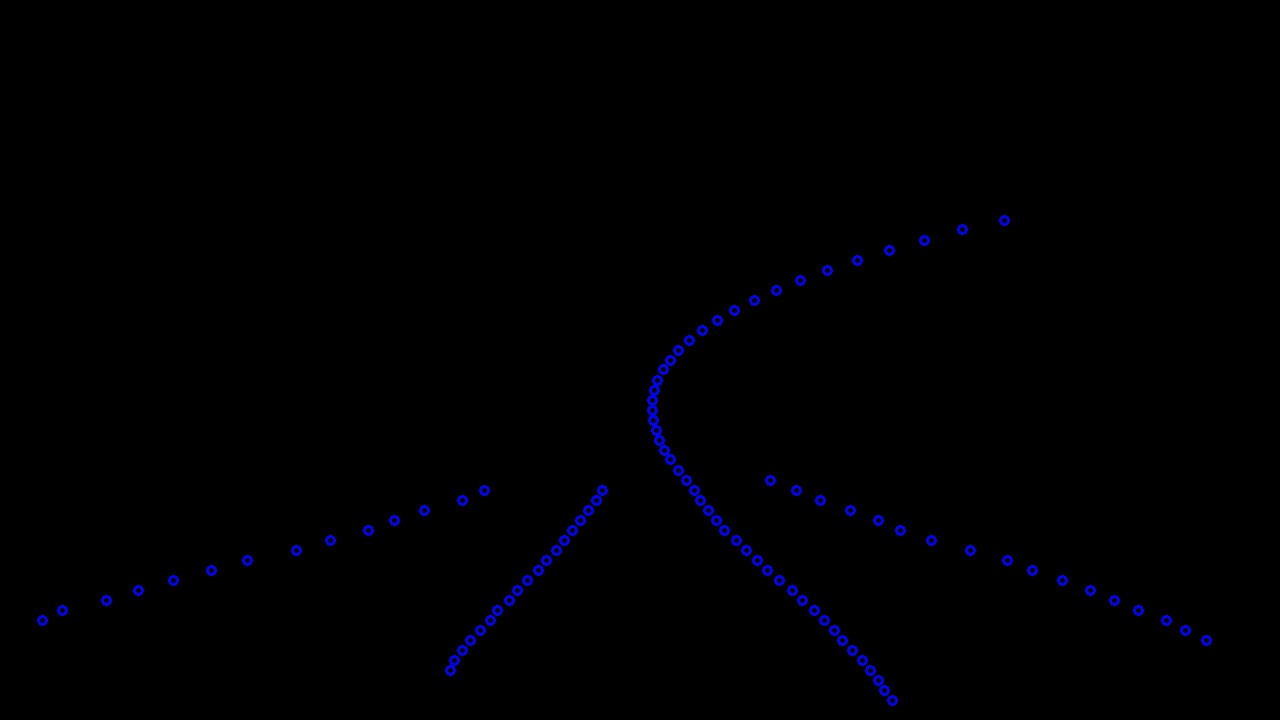}
     \end{subfigure}
     \begin{subfigure}[b]{0.20\textwidth}
         \includegraphics[width=\textwidth, height=\myheight cm]{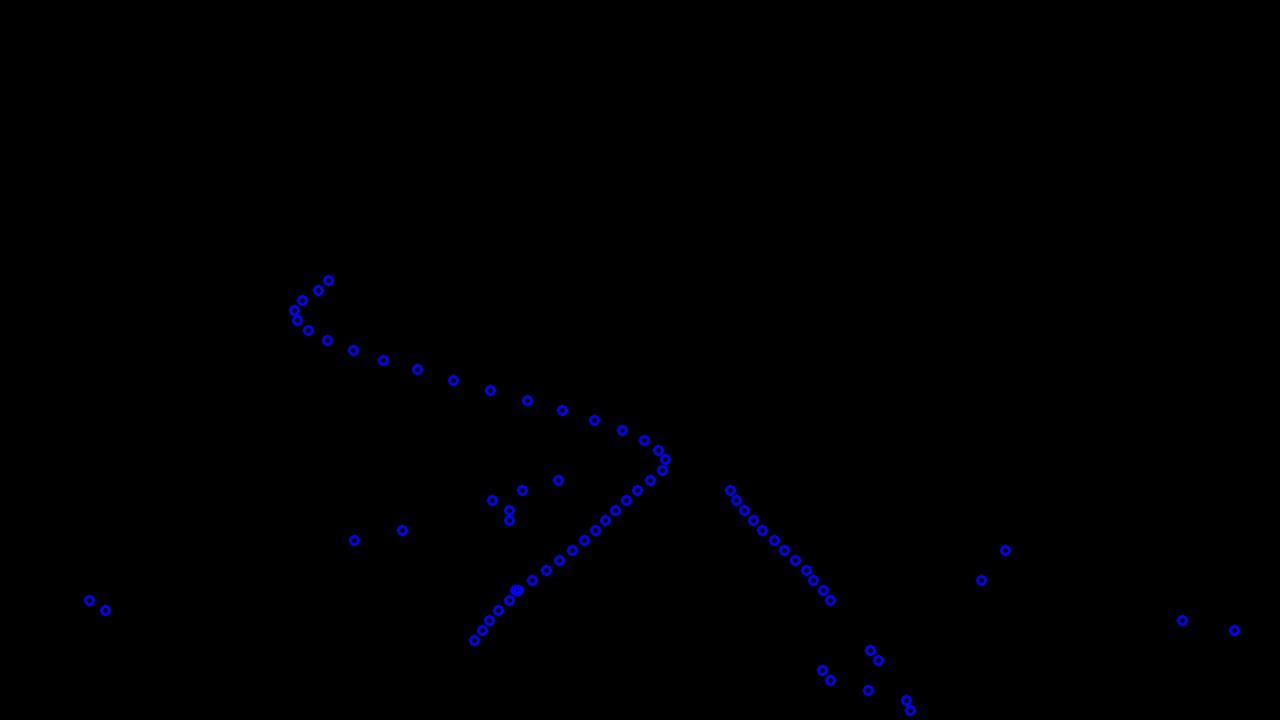}
     \end{subfigure}
     \begin{subfigure}[b]{0.19\textwidth}
         \includegraphics[width=\textwidth, height=\myheight cm]{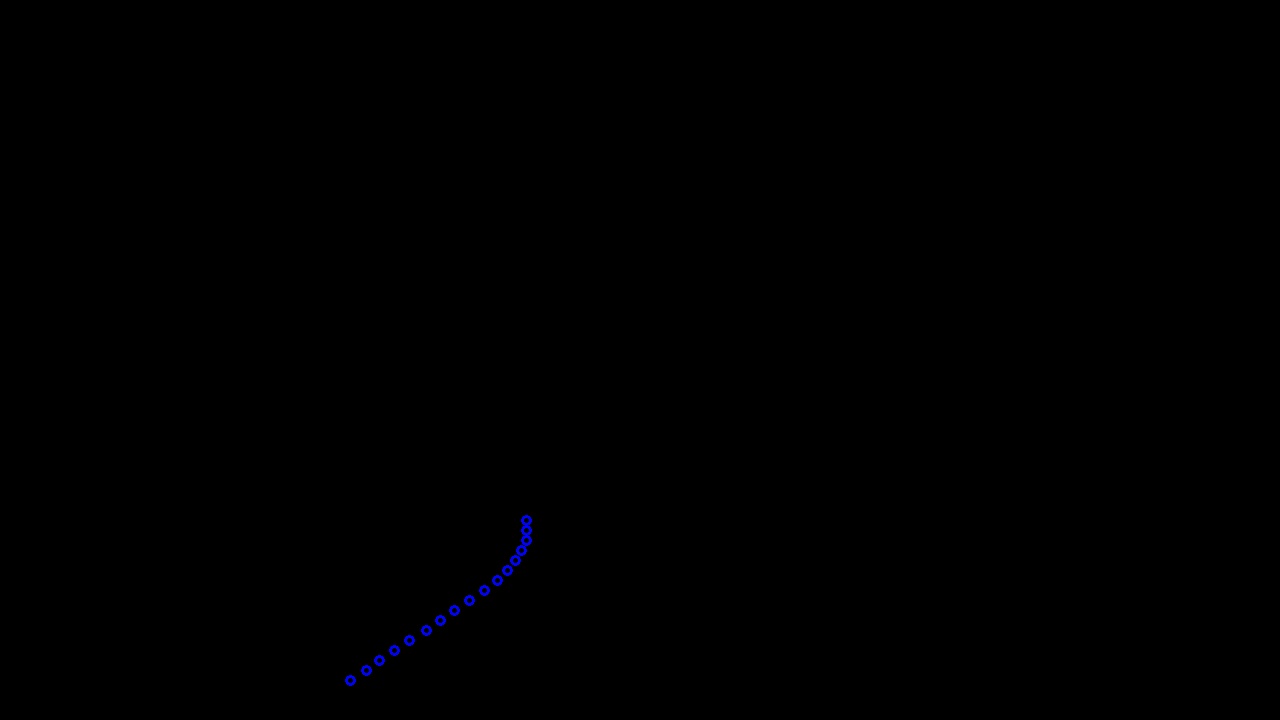}
     \end{subfigure}

     %\vspace{0.2cm}
     %\begin{subfigure}[b]{0.19\textwidth}
     %    \includegraphics[width=\textwidth, height=\myheight cm]{images/det-images/our-det1.jpg}
     %\end{subfigure}
     %\begin{subfigure}[b]{0.20\textwidth}
     %    \includegraphics[width=\textwidth, height=\myheight cm]{images/det-images/our-det2.jpg}
     %\end{subfigure}
     %\begin{subfigure}[b]{0.20\textwidth}
     %    \includegraphics[width=\textwidth, height=\myheight cm]{images/det-images/our-det3.jpg}
     %\end{subfigure}
     %\begin{subfigure}[b]{0.20\textwidth}
     %    \includegraphics[width=\textwidth, height=\myheight cm]{images/det-images/our-det4.jpg}
     %\end{subfigure}
     %\begin{subfigure}[b]{0.19\textwidth}
     %    \includegraphics[width=\textwidth, height=\myheight cm]{images/det-images/our-det5.jpg}
     %\end{subfigure}

     \vspace{0.2cm}
     \begin{subfigure}[b]{0.19\textwidth}
         \includegraphics[width=\textwidth, height=\myheight cm]{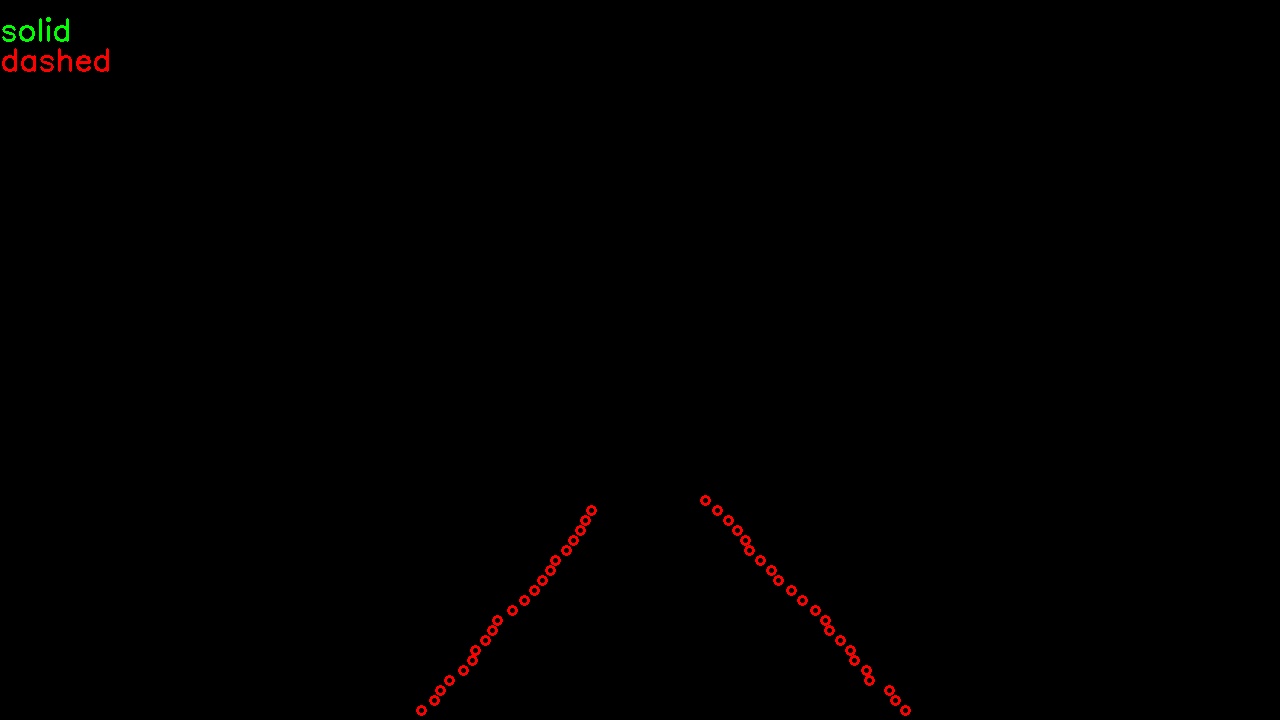}
     \end{subfigure}
     \begin{subfigure}[b]{0.20\textwidth}
         \includegraphics[width=\textwidth, height=\myheight cm]{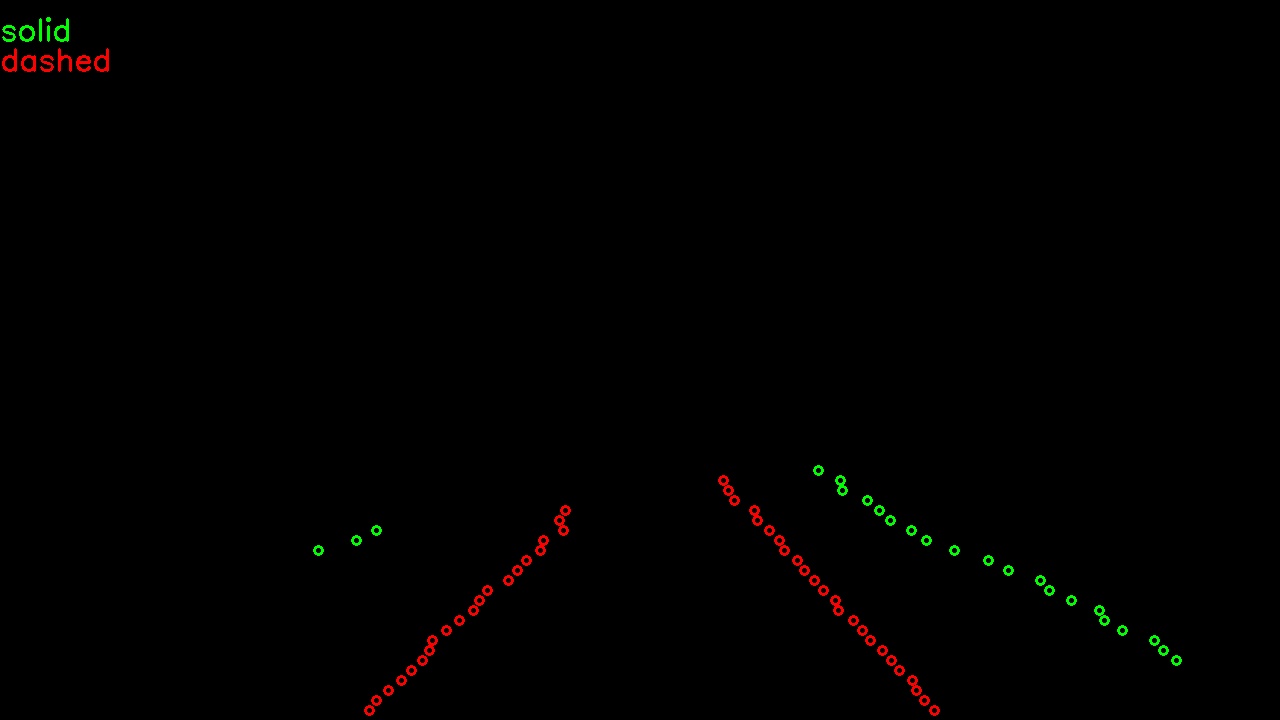}
     \end{subfigure}
     \begin{subfigure}[b]{0.20\textwidth}
         \includegraphics[width=\textwidth, height=\myheight cm]{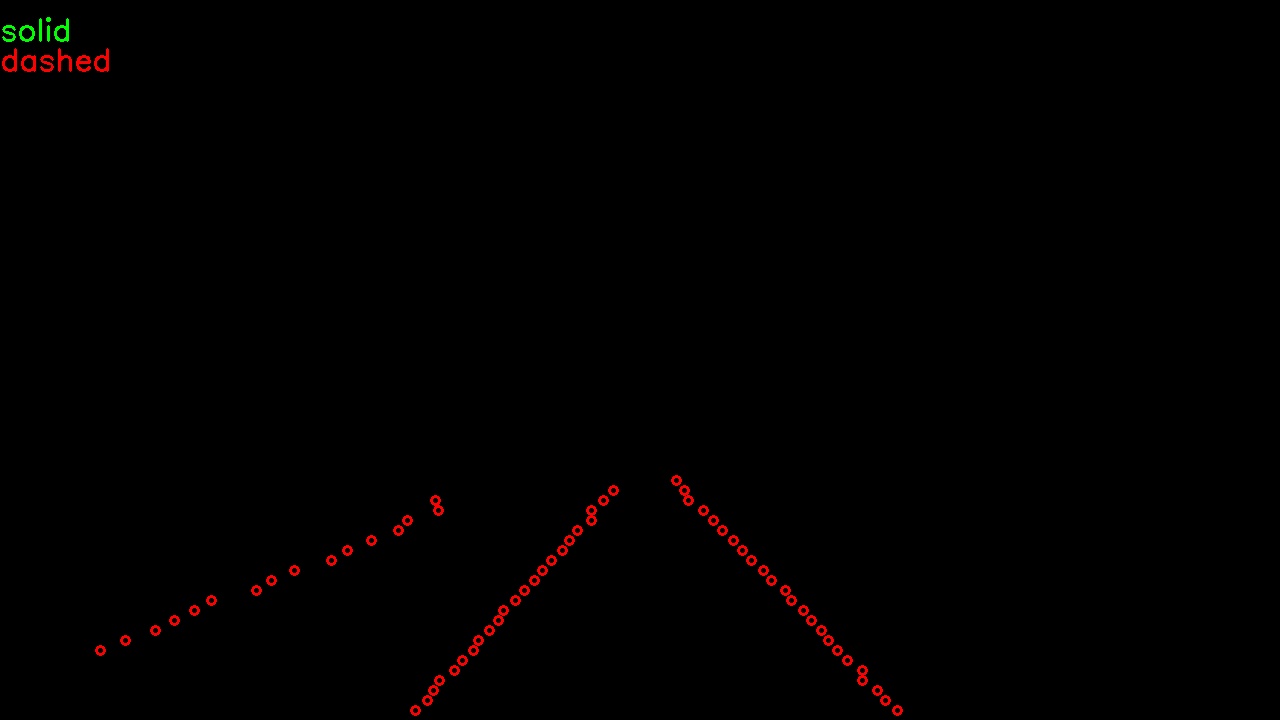}
     \end{subfigure}
     \begin{subfigure}[b]{0.20\textwidth}
         \includegraphics[width=\textwidth, height=\myheight cm]{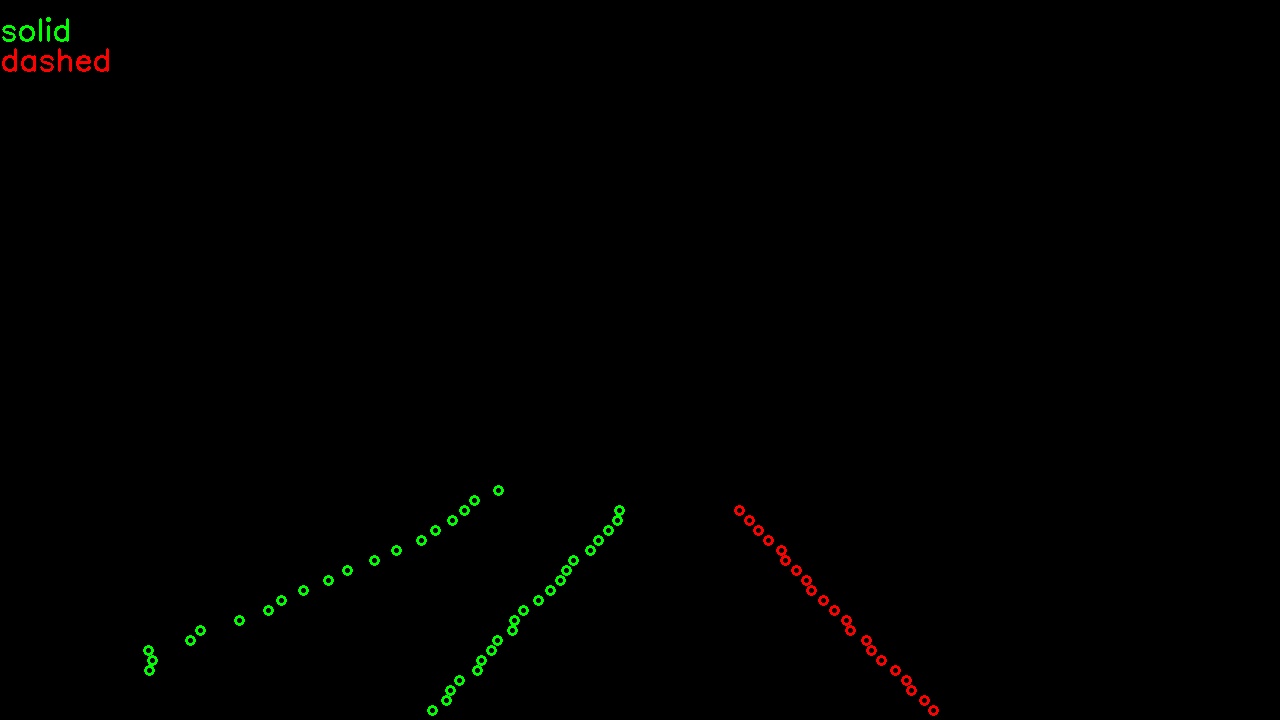}
     \end{subfigure}
     \begin{subfigure}[b]{0.19\textwidth}
         \includegraphics[width=\textwidth, height=\myheight cm]{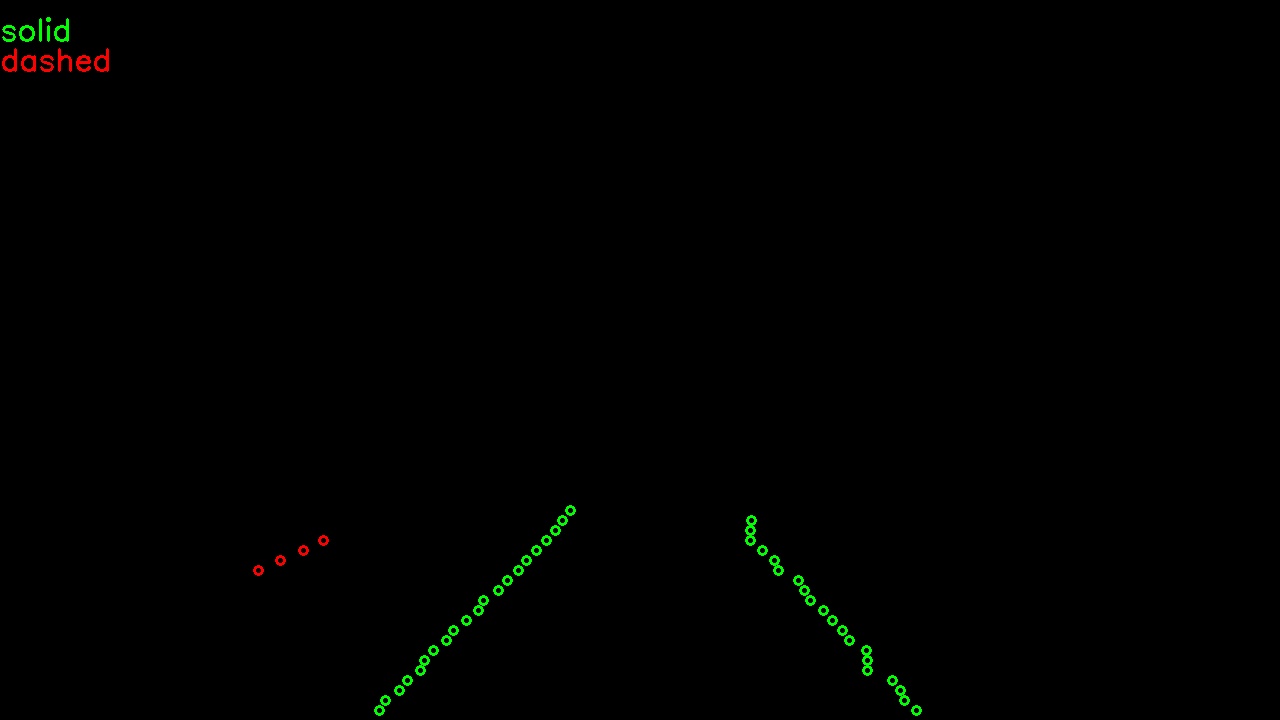}
     \end{subfigure}
 \caption{Detection performance of different models. From top to bottom: original, ground truth (solid - green, dashed - red), UFLD trained on TuSimple, RESA trained on TuSimple, UFLD trained on TuSimple+LVLane (our detection) + classification.}%, detection + classification}
 \label{fig: test-det-final}
\end{figure*}

We depict the performance of our model on the challenging LVLane test set in Fig. \ref{fig: test-det-final}, which includes scenarios with bright lighting conditions and partially visible lane markings. In these difficult cases, we observe a performance degradation in the UFLD and RESA models trained only on TuSimple, while our model maintains accurate lane localization. Furthermore, our model achieves correct lane classification for all instances. Additionally, through experimentation on various test images (not displayed here), we consistently observe superior detection and classification performance, particularly in challenging examples, validating the efficacy of our network.

Upon reviewing Table \ref{tab: result}, we observe the impact of mixed precision on the time performance of the models. Through a comprehensive analysis, we discover that when utilizing a large model such as RESA with a correspondingly large batch size to fully utilize the GPU's capacity, implementing mixed precision effectively reduces training and testing time without compromising performance. We present the training and inference time specifically for the RESA model as evidence of this improvement. Conversely, for lightweight models like UFLD, the application of mixed precision has minimal influence on the training and testing time.

We present a comparison of our proposed classification model with other existing methods in Table \ref{tab: comparison-result}. While only a limited number of existing systems have reported their evaluation performance, we provide our classification result on the TuSimple validation set for comparison. This model is trained only on the TuSimple train set for two classes (solid and dashed). The results clearly demonstrate the robust discriminating capability of our proposed system in distinguishing between solid and dashed types of lanes. These findings highlight the potential of our system to be utilized in ADAS solutions for effective lane change decision-making.
\begin{table}
\caption{Comparison of Class Accuracy on TuSimple}
\label{tab: comparison-result}
\begin{center}
\begin{tabular}{ |c|c|c| } 
 \hline
 \textbf{Method} & \textbf{2-class} \\ 
 \hline
 UDA \cite{sim} & 0.557\\ 
 Cascaded CNN \cite{pizzati2020lane} & 0.960\\ 
 Ours & \textbf{0.979}\\
 \hline
\end{tabular}
\end{center}
\end{table}

\subsection{Next Steps}
In order to bolster the learning capabilities of deep neural networks, we are expanding our dataset to include approximately 2500 images, carefully incorporating a wider range of variations.  This step is crucial as the existing 1000 images fall short in providing all the necessary environmental variability. %feature learning capacity. 
At present, the dataset comprises challenging instances with extreme lighting conditions, images featuring partially visible lane markings, and some cloudy weather images. Our future plans involve expanding the dataset to include rainy and cloudy images, as well as a wider range of white and yellow lane marking images. This expansion will enable the dataset to be used independently, eliminating the need for an additional TuSimple-like dataset. Further improvements include the incorporation of Transformers for classification and an additional free-space detection branch to handle scenarios lacking identifiable lane markings. 

\section{Conclusions}
In this paper, we present a deep-learning approach for lane detection and classification. We address the limitations of existing lane detection models, particularly in challenging scenarios. To overcome these limitations, we curate a comprehensive dataset and fine-tune the state-of-the-art models, leading to improved accuracy in lane localization. Additionally, we introduce a lightweight lane classifier that can be used with the detector to identify lane types, such as solid or dashed lines. These advancements have significant implications for lane-changing decisions in autonomous vehicles and ADAS systems. We further examine the mixed precision method's effectiveness and find it useful for large models and datasets. Our model is evaluated on public benchmarks of TuSimple and Caltech datasets and the newly curated LVLane dataset, demonstrating superior performance in localizing lanes in challenging scenarios and real-time classification of lane types.

\bibliographystyle{IEEEtran}
\bibliography{mybibfile}

% Generated by IEEEtran.bst, version: 1.14 (2015/08/26)
\begin{thebibliography}{10}
\providecommand{\url}[1]{#1}
\csname url@samestyle\endcsname
\providecommand{\newblock}{\relax}
\providecommand{\bibinfo}[2]{#2}
\providecommand{\BIBentrySTDinterwordspacing}{\spaceskip=0pt\relax}
\providecommand{\BIBentryALTinterwordstretchfactor}{4}
\providecommand{\BIBentryALTinterwordspacing}{\spaceskip=\fontdimen2\font plus
\BIBentryALTinterwordstretchfactor\fontdimen3\font minus
  \fontdimen4\font\relax}
\providecommand{\BIBforeignlanguage}[2]{{%
\expandafter\ifx\csname l@#1\endcsname\relax
\typeout{** WARNING: IEEEtran.bst: No hyphenation pattern has been}%
\typeout{** loaded for the language `#1'. Using the pattern for}%
\typeout{** the default language instead.}%
\else
\language=\csname l@#1\endcsname
\fi
#2}}
\providecommand{\BIBdecl}{\relax}
\BIBdecl

\bibitem{bar2014recent}
A.~Bar~Hillel, R.~Lerner, D.~Levi, and G.~Raz, ``Recent progress in road and
  lane detection: a survey,'' \emph{Machine vision and applications}, vol.~25,
  no.~3, pp. 727--745, 2014.

\bibitem{pan2018spatial}
X.~Pan, J.~Shi, P.~Luo, X.~Wang, and X.~Tang, ``Spatial as deep: Spatial cnn
  for traffic scene understanding,'' in \emph{Proceedings of the AAAI
  Conference on Artificial Intelligence}, vol.~32, no.~1, 2018.

\bibitem{qin2020ultra}
Z.~Qin, H.~Wang, and X.~Li, ``Ultra fast structure-aware deep lane detection,''
  in \emph{Computer Vision--ECCV 2020: 16th European Conference, Glasgow, UK,
  August 23--28, 2020, Proceedings, Part XXIV 16}.\hskip 1em plus 0.5em minus
  0.4em\relax Springer, 2020, pp. 276--291.

\bibitem{zheng2021resa}
T.~Zheng, H.~Fang, Y.~Zhang, W.~Tang, Z.~Yang, H.~Liu, and D.~Cai, ``Resa:
  Recurrent feature-shift aggregator for lane detection,'' in \emph{Proceedings
  of the AAAI Conference on Artificial Intelligence}, vol.~35, no.~4, 2021, pp.
  3547--3554.

\bibitem{zheng2022clrnet}
T.~Zheng, Y.~Huang, Y.~Liu, W.~Tang, Z.~Yang, D.~Cai, and X.~He, ``Clrnet:
  Cross layer refinement network for lane detection,'' in \emph{Proceedings of
  the IEEE/CVF conference on computer vision and pattern recognition}, 2022,
  pp. 898--907.

\bibitem{tusimple}
``{TuSimple. Tusimple benchmark},
  {\url{https://github.com/tusimple/tusimple-benchmark/}}, {Accessed:
  2023-01-05}.''

\bibitem{aly2008real}
M.~Aly, ``Real time detection of lane markers in urban streets,'' in \emph{2008
  IEEE intelligent vehicles symposium}.\hskip 1em plus 0.5em minus 0.4em\relax
  IEEE, 2008, pp. 7--12.

\bibitem{micikevicius2017mixed}
P.~Micikevicius, S.~Narang, J.~Alben, G.~Diamos, E.~Elsen, D.~Garcia,
  B.~Ginsburg, M.~Houston, O.~Kuchaiev, G.~Venkatesh \emph{et~al.}, ``Mixed
  precision training,'' \emph{arXiv preprint arXiv:1710.03740}, 2017.

\bibitem{sun2006hsi}
T.-Y. Sun, S.-J. Tsai, and V.~Chan, ``Hsi color model based lane-marking
  detection,'' in \emph{2006 ieee intelligent transportation systems
  conference}.\hskip 1em plus 0.5em minus 0.4em\relax IEEE, 2006, pp.
  1168--1172.

\bibitem{wang2000lane}
Y.~Wang, D.~Shen, and E.~K. Teoh, ``Lane detection using spline model,''
  \emph{Pattern Recognition Letters}, vol.~21, no.~8, pp. 677--689, 2000.

\bibitem{rahman2020real}
Z.~Rahman, A.~M. Ami, and M.~A. Ullah, ``A real-time wrong-way vehicle
  detection based on yolo and centroid tracking,'' in \emph{2020 IEEE Region 10
  Symposium (TENSYMP)}.\hskip 1em plus 0.5em minus 0.4em\relax IEEE, 2020, pp.
  916--920.

\bibitem{xu2020curvelane}
H.~Xu, S.~Wang, X.~Cai, W.~Zhang, X.~Liang, and Z.~Li, ``Curvelane-nas:
  Unifying lane-sensitive architecture search and adaptive point blending,'' in
  \emph{Computer Vision--ECCV 2020: 16th European Conference, Glasgow, UK,
  August 23--28, 2020, Proceedings, Part XV 16}.\hskip 1em plus 0.5em minus
  0.4em\relax Springer, 2020, pp. 689--704.

\bibitem{liu2021condlanenet}
L.~Liu, X.~Chen, S.~Zhu, and P.~Tan, ``Condlanenet: a top-to-down lane
  detection framework based on conditional convolution,'' in \emph{Proceedings
  of the IEEE/CVF international conference on computer vision}, 2021, pp.
  3773--3782.

\bibitem{de2015automatic}
M.~B. de~Paula and C.~R. Jung, ``Automatic detection and classification of road
  lane markings using onboard vehicular cameras,'' \emph{IEEE Transactions on
  Intelligent Transportation Systems}, vol.~16, no.~6, pp. 3160--3169, 2015.

\bibitem{song2018lane}
W.~Song, Y.~Yang, M.~Fu, Y.~Li, and M.~Wang, ``Lane detection and
  classification for forward collision warning system based on stereo vision,''
  \emph{IEEE Sensors Journal}, vol.~18, no.~12, pp. 5151--5163, 2018.

\bibitem{pizzati2020lane}
F.~Pizzati, M.~Allodi, A.~Barrera, and F.~Garc{\'\i}a, ``Lane detection and
  classification using cascaded cnns,'' in \emph{Computer Aided Systems
  Theory--EUROCAST 2019: 17th International Conference, Las Palmas de Gran
  Canaria, Spain, February 17--22, 2019, Revised Selected Papers, Part II
  17}.\hskip 1em plus 0.5em minus 0.4em\relax Springer, 2020, pp. 95--103.

\bibitem{sim}
C.~Hu, S.~Hudson, M.~Ethier, M.~Al-Sharman, D.~Rayside, and W.~Melek,
  ``Sim-to-real domain adaptation for lane detection and classification in
  autonomous driving,'' in \emph{2022 IEEE Intelligent Vehicles Symposium
  (IV)}, 2022, pp. 457--463.

\bibitem{dosovitskiy2017carla}
A.~Dosovitskiy, G.~Ros, F.~Codevilla, A.~Lopez, and V.~Koltun, ``Carla: An open
  urban driving simulator,'' in \emph{Conference on robot learning}.\hskip 1em
  plus 0.5em minus 0.4em\relax PMLR, 2017, pp. 1--16.

\bibitem{dutta2016via}
A.~Dutta, A.~Gupta, and A.~Zissermann, ``{VGG} image annotator ({VIA}),''
  http://www.robots.ox.ac.uk/~vgg/software/via/, 2016, version: X.Y.Z,
  Accessed: 2023-01-10.

\bibitem{dutta2019vgg}
\BIBentryALTinterwordspacing
A.~Dutta and A.~Zisserman, ``The {VIA} annotation software for images, audio
  and video,'' in \emph{Proceedings of the 27th ACM International Conference on
  Multimedia}, ser. MM '19.\hskip 1em plus 0.5em minus 0.4em\relax New York,
  NY, USA: ACM, 2019. [Online]. Available:
  \url{https://doi.org/10.1145/3343031.3350535}
\BIBentrySTDinterwordspacing

\bibitem{he2016deep}
K.~He, X.~Zhang, S.~Ren, and J.~Sun, ``Deep residual learning for image
  recognition,'' in \emph{Proceedings of the IEEE conference on computer vision
  and pattern recognition}, 2016, pp. 770--778.

\bibitem{paszke2019pytorch}
A.~Paszke, S.~Gross, F.~Massa, A.~Lerer, J.~Bradbury, G.~Chanan, T.~Killeen,
  Z.~Lin, N.~Gimelshein, L.~Antiga \emph{et~al.}, ``Pytorch: An imperative
  style, high-performance deep learning library,'' \emph{Advances in neural
  information processing systems}, vol.~32, 2019.

\end{thebibliography}

\end{document}